\documentclass[10pt,twocolumn,letterpaper]{article}

\usepackage[pagenumbers]{cvpr} 

\usepackage[T1]{fontenc}
\usepackage{graphicx}
\usepackage{amsmath,amssymb}
\usepackage{tabularray}
\usepackage{booktabs,tabularx,multirow}
\usepackage{xcolor,xspace}
\usepackage{microtype}
\usepackage{xcolor}
\usepackage{makecell}
\usepackage{stfloats}

\newcommand{\edittype}[1]{%
  \begingroup
  \setlength{\fboxsep}{0pt}%
  \colorbox{gray!24}{%
    \rule[-0.5ex]{0pt}{2.5ex}%
    \text{#1}%
  }%
  \endgroup
}

\newcommand{\method}{EditAgent\xspace}

\definecolor{lightblue}{rgb}{0.2, 0.5, 0.9}

\usepackage{listings}
\usepackage{algorithm}
\usepackage{algpseudocode}
\usepackage{needspace}
\usepackage{placeins}
\usepackage{array}
\newcolumntype{Y}{>{\raggedright\arraybackslash}X}
\newcolumntype{P}[1]{>{\raggedright\arraybackslash}p{#1}}

\input{glyphtounicode}
\lstdefinestyle{prompt}{
  basicstyle=\ttfamily\fontsize{7.5}{8.7}\selectfont,
  breaklines=true, breakatwhitespace=true, columns=fullflexible, keepspaces=true,
  frame=single, framerule=0.3pt, rulecolor=\color{gray!45}, backgroundcolor=\color{gray!4},
  framesep=3pt, xleftmargin=3pt, xrightmargin=3pt, aboveskip=4pt, belowskip=8pt,
  showstringspaces=false, upquote=true, extendedchars=true, escapeinside={(*@}{@*)},
  breakindent=0pt, breakautoindent=true,
  literate={—}{{\textemdash}}1 {~}{{$\sim$\nobreak}}1 {★}{{$\star$}}1 {°}{{$^{\circ}$}}1 {→}{{$\rightarrow$}}1
           {…}{{\ldots}}1 {í}{{\'i}}1 {×}{{$\times$}}1 {≤}{{$\le$}}1 {≥}{{$\ge$}}1
           {±}{{$\pm$}}1 {’}{{'}}1 {“}{{``}}1 {”}{{''}}1
           {⟨}{{$\langle$}}1 {⟩}{{$\rangle$}}1
}
\newcommand{\promptlisting}[4][10]{\Needspace{#1\baselineskip}%
  \lstinputlisting[style=prompt, caption={#3}, label={#4}]{#2}}

\algrenewcommand\algorithmiccomment[1]{\hfill{\color{gray}\(\triangleright\) #1}}

\definecolor{cvprblue}{rgb}{0.21,0.49,0.74}
\usepackage[pagebackref,breaklinks,colorlinks,allcolors=cvprblue]{hyperref}

\def\paperID{*****} 
\def\confName{CVPR}
\def\confYear{2026}

\title{\method: Rethinking Agentic Image Editing}

\author{
    Haiyang Xu$^{1*}$
    \hspace{0.7cm} Tianyu Wang$^2$
    \hspace{0.7cm} Jialu Li$^2$
    \hspace{0.7cm} Jiteng Mu$^2$ \\
    \hspace{0.7cm} Yuanjun Xiong$^2$
    \hspace{0.7cm} Jianming Zhang$^2$
    \hspace{0.7cm} Zhuowen Tu$^1$
    \hspace{0.7cm} Mingze Xu$^2$ \\ [.5ex]
    $^1$UC San Diego \hspace{0.9cm} $^2$Adobe Research \\ [.5ex]
    {\tt\small hax038@ucsd.edu; mingzex@adobe.com}
}

\begin{document}


\twocolumn[{
    \maketitle
    \vspace{-3em}
    \begin{center}
    \captionsetup{type=figure}
    \includegraphics[width=\linewidth]{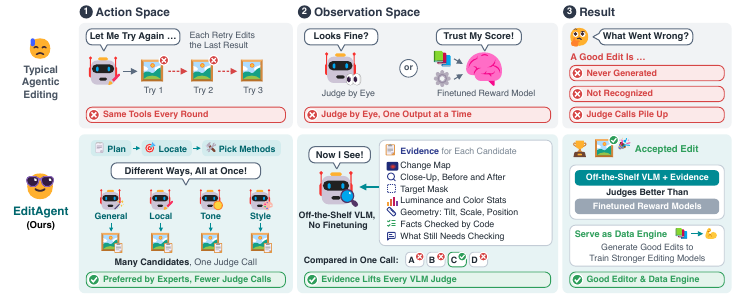}
    \vspace{-2em}
    \captionof{figure}{\textbf{\method expands both the action space of the agent and the observation space of its judge.} Instead of trying again with the same editor, \method generates candidates with different editing methods in parallel, and a judge (off-the-shelf VLM) compares them in one call with a verification record (evidence) for each candidate, without judge finetuning.}
    \label{fig:teaser}
    \end{center}
}]

{\renewcommand{\thefootnote}{\fnsymbol{footnote}}
\footnotetext[1]{Work done during an internship at Adobe Research.}}

\begin{abstract}
Agentic image editors decompose complex requests into steps, but when a step fails, they mostly retry with the same narrow set of tools and a judge that scores each output by eye directly. We argue that such agents are bounded by two spaces: the action space of what the agent can do, and the observation space of what its judge can see. On our expert-annotated benchmark, a general-purpose editor yields no usable result in over 40\% of the cases, and no judge is reliable across edit types. We present \method, which expands both spaces. It routes each step to general editors and specialized methods, and attaches to each candidate a verification record. With these records, off-the-shelf VLM judges outperform finetuned reward models. Candidates are generated in parallel and judged in one listwise call: experts prefer this to sequential refinement in 78.5\% of decided comparisons, at no more than half the judge calls. \method improves two open-weight editors on three benchmarks and serves as a data engine for training stronger editors.
\end{abstract}

\vspace{-2em}
\section{Introduction}
\label{sec:introduction}

Instruction-based image editing models can now perform a wide range of edits from a single text instruction~\cite{brooks2023instructpix2pix,wu2025qwenimagetechnicalreport,flux-2-2025}. Real requests, however, rarely ask for a single edit. Given ``move me to the left of the frame, widen the shot to show more of the beach, remove the people walking behind me, and make it look like sunset,'' today's editors tend to complete some parts and silently drop others~\cite{yang2025complexedit,wang2025complexbenchedit}. Agentic editing systems~\cite{gupta2025costa,liang2025refineedit,shen2026imagagent,zeng2026mira} address this by decomposing the request into steps and running a plan-edit-judge loop on each. Yet when a step fails, they mostly try again. The step is refined, retried, or rolled back with the same tools and judge. Here we ask the question: \textbf{when an editing agent fails, is another attempt really what it needs? If not, what does the agent really needs?}

Consider how a professional retoucher would handle the same request: pick a different tool for each part, and check each result by zooming in and comparing before and after. A retoucher does not simply try again but acts differently and looks more closely. We argue that an editing agent is likewise bounded by two spaces: its \textbf{action space}, what it can do, and its judge's \textbf{observation space}, what it can see.
Existing agents~\cite{gupta2025costa,liang2025refineedit,shen2026imagagent} keep both small. Each attempt runs one tool, often a general editor~\cite{bfl2025kontext,flux-2-2025,wu2025qwenimagetechnicalreport,openai2026gptimage25,google2025geminiimage}, and the judge scores that single output by eye, given only the source image, the edited image and the instruction.
To explore how each space should be designed, we introduce a diagnostic benchmark of 600 candidate sets over 30 edit types, labeled blindly by professional Photoshop editors (\cref{fig:motivation}). We find that the general editor yields no usable result in 42.7\% of the sets, mostly in geometry, tone and enhancement edits, and a specialized method recovers 175 of these 256 failures. Meanwhile, no judge we test, including finetuned reward models~\cite{luo2026editscore,wu2026editreward}, is the most accurate across edit groups.

We therefore rethink agentic editing by expanding both spaces. Our system, \textbf{\method}, shown in~\cref{fig:teaser}, expands the action space by routing each step to general-purpose editors and specialized methods, including masked local editing, object removal and movement, photometric and geometric operators, and so on. It expands the observation space by attaching to each candidate a verification record (evidence), such as a change map, close-ups of the edited region, and measured color and geometry, together with the outcomes of programmatic checks, so that the judge knows what code has verified and what remains for visual judgment. Instead of refining one result round after round, \method thus generates candidates from all selected methods in parallel from the same source image, and the judge compares them with their records in a single listwise call. Unlike in text-to-image generation~\cite{jaiswal2026iterative}, professional editors prefer this parallel search over sequential refinement in 78.5\% of non-tied comparisons, with at most half the judge calls (\cref{sec:search}). With verification records, off-the-shelf VLM judges outperform finetuned reward models without finetuning.

Instantiated with Qwen-Image-Edit-2511~\cite{wu2025qwenimagetechnicalreport} and FLUX.2-klein-9B~\cite{bfl2026flux2klein}, \method improves both backbones on EdiVal~\cite{chen2025edival}, Complex-Edit~\cite{yang2025complexedit}, and ComplexBench-Edit~\cite{wang2025complexbenchedit}. Because it accepts only verified edits, \method also serves as a data engine. Editors finetuned on its 25.7K accepted sessions outperform those trained on data generated directly by the base editor FLUX.2-klein-9B.

Our contributions are threefold.
\begin{itemize}[leftmargin=*, labelindent=\parindent, itemsep=1pt, topsep=2pt]
    \item We identify the agent's action space and its judge's observation space as two bottlenecks of agentic image editing, and analyze both with an expert-annotated diagnostic benchmark spanning 30 edit types.
    \item We propose \method, which expands the action space with specialized editing methods and the observation space with verification records, and couples them through parallel candidate generation and listwise judging. With these records, off-the-shelf VLM judges outperform finetuned editing reward models.
    \item \method consistently improves strong open-weight editors on three multi-turn and complex-instruction benchmarks, and the sessions it accepts yield training data that further improves standalone editors.
\end{itemize}

\vspace{-0.25em}
\section{Related Work}
\vspace{-0.25em}
\label{sec:related}

\noindent \textbf{Single-turn Image Editing.}
Following pioneering diffusion-based methods~\cite{meng2022sdedit,hertz2022prompt,mokady2023nulltext,kawar2023imagic}, InstructPix2Pix~\cite{brooks2023instructpix2pix} and Emu Edit~\cite{sheynin2024emu} established general instruction-based editing.
Recent proprietary~\cite{openai2026gptimage25,google2025geminiimage,wang2025seededit3,seedream2025seedream4} and open-weight~\cite{liu2025step1x,wu2025qwenimagetechnicalreport,bfl2025kontext,bfl2026flux2klein,deng2025bagel,wu2025omnigen2} models now provide strong general-purpose editing.
Other methods offer more explicit control through masked editing~\cite{suvorov2022lama,avrahami2022blended}, localized text-driven editing~\cite{couairon2022diffedit,li2024zone}, or point-based manipulation~\cite{shi2024dragdiffusion}. \method builds on some of these editors and coordinates them together.

\noindent \textbf{Multi-turn Editing and Agentic Editing.}
MagicBrush~\cite{zhang2023magicbrush} provides single- and multi-turn editing trajectories, making the preservation of earlier edits an explicit target.
Agentic systems further incorporate planning and feedback, for example by scheduling restoration tools~\cite{zhu2025agenticir}, searching and revising cost-aware tool paths~\cite{gupta2025costa}, and guiding iterative edits with instruction decomposition, visual feedback, or dialogue history~\cite{liang2025refineedit,ma2025talk2image,shen2026imagagent,zeng2026mira,wang2024genartist,li2025editthinker,yang2026riseagent,chu2026visiondirector}.
Others integrate reflection into the editing model~\cite{yin2025reasonedit}, maintain a structured history for long-horizon editing~\cite{ye2026agentbanana}, or explore editing trajectories with Monte Carlo tree search~\cite{yao2026photoagent}.
These agents retry within a fixed tool set and judge one output at a time by eye, whereas \method expands the action space with specialized editing methods and the observation space of its judge with evidence and side-by-side comparison.

\noindent \textbf{Evaluation and Benchmarking.}
Traditional protocols measure edit alignment with CLIP-based text-image or directional similarity~\cite{radford2021clip,gal2022stylegan} and preservation with CLIP or DINO image similarity~\cite{caron2021dino}.
Benchmarks further compare outputs with ground-truth target images~\cite{zhang2023magicbrush} or assess unedited regions~\cite{ju2024directinversion} with PSNR, SSIM~\cite{wang2004ssim}, and LPIPS~\cite{zhang2018lpips}.
More recent protocols prompt VLMs such as GPT-4o~\cite{openai2024gpt4o} and Qwen2.5-VL~\cite{bai2025qwen25vl} to rate instruction following, preservation, and perceptual quality~\cite{ku2024viescore,liu2025step1x,yang2025complexedit,ye2025imgedit}.
Others augment VLM judgments with visual tools~\cite{chen2025edival,wang2026i2ibench} or aggregate automatic scores into a human-aligned evaluation~\cite{ryu2025hatie}.
Rather than training dedicated editing reward models~\cite{zhang2024hive,luo2026editscore,wu2026editreward}, we adapt off-the-shelf VLMs with task-specific verification records.

\vspace{-0.25em}
\section{Rethinking Agentic Editing}
\vspace{-0.25em}
\label{sec:rethinking}

\begin{figure*}[!th]
    \centering
    \includegraphics[width=\linewidth]{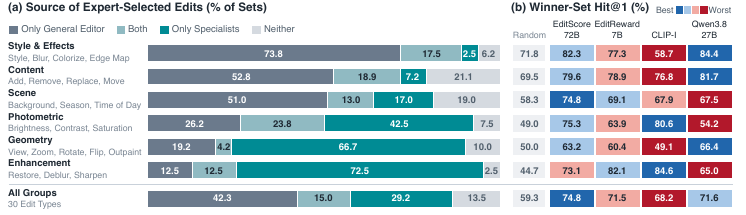}
    \vspace{-2em}
    \caption{\textbf{Where the action space and the observation space fall short, by edit group.} 
    (a) Action space: share of the expert-labeled sets in which only general-editor candidates, only specialized candidates, both kinds, or no candidate are usable. (b) Observation space: Winner-Set Hit@1 on the expert-labeled sets with at least one usable candidate.
    Random is the expected Hit@1 of a uniform random pick.
    }
    \vspace{-1em}
    \label{fig:motivation}
\end{figure*}

We evaluate both action and observation spaces on the diagnostic benchmark (\cref{sec:diagnostic_benchmark}, \cref{fig:motivation}). Specifically, we address three core questions: (1) Does a general editor need a larger action space or only more samples (\cref{sec:missing_candidate})? (2) Under a fixed budget, should candidates be refined sequentially or generated in parallel (\cref{sec:search})? (3) Can a judge that scores one output at a time select a satisfactory edit (\cref{sec:missed_candidate})?

\subsection{Diagnostic Benchmark}
\label{sec:diagnostic_benchmark}

We construct a benchmark of 600 editing sets covering 30 fine-grained edit types, with 20 sets per type. The types span six groups: \edittype{Content}, \edittype{Style \& Effects}, \edittype{Scene}, \edittype{Photometric}, \edittype{Geometry}, and \edittype{Enhancement}. Each set contains a source image, an instruction, and 2-4 candidate edits from two sources: samples of a general-purpose editor (FLUX.2-klein-9B) and specialized methods such as masked local editing, object removal, and photometric operators. Five professional Photoshop retouchers sit together and select every usable candidate, \textit{i.e.}, every candidate that completes the instruction acceptably. We call the selected candidates the winner set.

\vspace{-0.5em}
\subsection{Is One Editor Enough?}
\vspace{-0.5em}

\label{sec:missing_candidate}

We categorize the 600 sets by their winner sets (\cref{fig:motivation}a). The winner set contains only general-editor candidates in 254 sets (42.3\%), only specialized candidates in 175 (29.2\%), and both kinds in 90 (15.0\%). In the remaining 81 sets (13.5\%), no candidate is usable. Since every set includes a general-editor candidate, the general editor fails in 256 sets (42.7\%), and a specialized method recovers 175 of them. The outcomes also vary strongly across edit groups. The winner set contains only general-editor candidates in 73.8\% of \edittype{Style \& Effects} sets, but only specialized candidates in 72.5\% of \edittype{Enhancement} and 66.7\% of \edittype{Geometry} sets.

More samples from the general editor do not recover these failures. For every request, the general editor produces several samples and one of them is delivered, \textit{i.e.}, parallel sampling with a verifier~\cite{jaiswal2026iterative}. A second pool also includes the specialized methods, and the same VLM judge selects the delivered edit in both pools. Both delivered edits are among the labeled candidates of the set. Although the general-editor pool contains more samples, its delivered edit is usable in only 45.2\% of the sets, compared with 68.2\% for the pool with specialized methods. In the 296 sets where exactly one delivered edit is usable, it comes from the pool with specialized methods in 217. \textbf{These results show that one editor is not enough, even when sampled more often.}

\subsection{Refine Sequentially or Search in Parallel?}
\vspace{-0.33em}
\label{sec:search}

While sequential refinement has been shown to improve text-image alignment in image generation~\cite{jaiswal2026iterative}, we observe that it can suffer from error accumulation in image editing, as errors introduced in earlier iterations may propagate to later edits. To investigate this, we compare two strategies using the same editor (FLUX.2-klein-9B) and judge (Qwen3.8-27B), each with a budget of 4 image generations. Sequential refinement iteratively edits the previous result using prompts revised based on VLM feedback, requiring 3 feedback calls and 1 final selection call. Parallel search, in contrast, generates two candidates independently from the original image. If neither is accepted by the judge, it revises the prompt and generates two more candidates from the original image before selecting the best result. This requires at most 2 VLM calls.

In a blind comparison on 144 Complex-Edit samples~\cite{yang2025complexedit} at complexity levels 1, 4, and 8, five professional Photoshop editors preferred parallel search in 102 cases and sequential refinement in 28, with 14 ties. Parallel search thus wins 78.5\% of the non-tied comparisons with at most half the VLM calls. Unlike generation, editing must stay faithful to the source image. Sequential refinement keeps editing its own output, so small errors can accumulate, whereas parallel search edits every candidate directly from the source.
\textbf{Thus, for editing, parallel search uses the budget better than sequential refinement.} With several candidates available, the delivered edit then depends on what the judge can see.

\begin{figure*}[t]
    \centering
    \includegraphics[width=\textwidth]{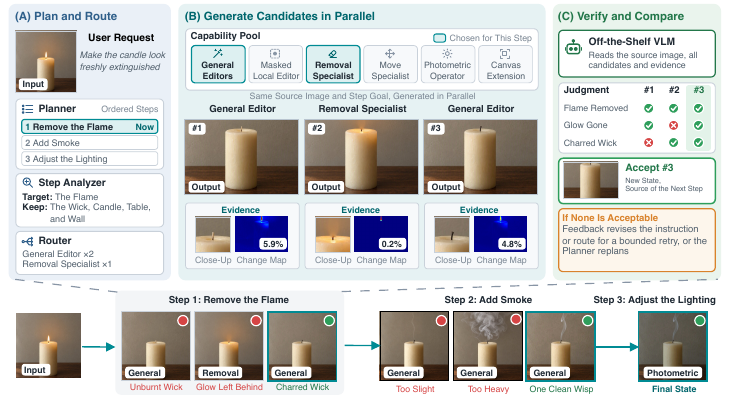}
    \vspace{-2em}
    \caption{
        \textbf{Overview of \method.}
        Bottom: a request runs as dependent steps, and each accepted result becomes the source of the next step.
        Top: step~1, enlarged. 
        (A) The planner orders the steps; the analyzer and router select the applicable methods for the current step. 
        (B) The selected methods generate candidates in parallel from the same source image, each with a verification record (evidence). 
        (C) An off-the-shelf VLM compares the candidates with their records and accepts one, or returns feedback for another attempt.
    }
    \vspace{-1em}
    \label{fig:system}
\end{figure*}

\vspace{-0.33em}
\subsection{Is Judging One Output at a Time Enough?}
\vspace{-0.33em}
\label{sec:missed_candidate}

We therefore test four judges, each scores one output at a time from the source image, the edited image and, except for CLIP-I, the instruction. They are the editing reward models EditScore-72B~\cite{luo2026editscore} and EditReward-7B~\cite{wu2026editreward}, the CLIP-I metric~\cite{radford2021clip} (CLIP similarity to the source image), and Qwen3.8-27B~\cite{qwen38} with a pass/fail prompt. Winner-Set Hit@1 is the fraction of sets whose top-scoring candidate is in the winner set, \textit{i.e.}, is usable, with ties sharing credit. We report it on the 515 sets with at least one usable candidate.

\begin{table}[t]
\centering
\caption{\textbf{Comparative judging is better.} Overall accuracy (\%) on EditReward-Bench~\cite{luo2026editscore}. Pointwise scores each candidate alone; listwise (ours) compares them together in one call.}
\label{tab:vlm-vs-rm}
\vspace{-0.75em}
\footnotesize
\setlength{\tabcolsep}{6pt}
\begin{tabular}{@{}lccc@{}}
\toprule
\textbf{Judge} & \textbf{Pointwise} & \textbf{Listwise (Ours)} & $\boldsymbol{\Delta}$ \\
\midrule
EditScore-7B & 65.9 & -- & -- \\
EditScore-32B & 68.0 & -- & -- \\
EditScore-72B & 70.3 & -- & -- \\
EditScore-7B, Ensemble@4 & 72.7 & -- & -- \\
EditScore-32B, Ensemble@4 & 73.3 & -- & -- \\
EditScore-72B, Ensemble@4 & 76.3 & -- & -- \\
\midrule
Qwen3.5-27B & 71.7 & 85.2 & +13.5 \\
Qwen3.8-27B & 72.1 & 85.9 & +13.8 \\
Gemma-4-31B & 73.8 & \textbf{86.1} & +12.3 \\
\bottomrule
\end{tabular}
\vspace{-2.25em}
\end{table}

No judge is the most accurate on every group (\cref{fig:motivation}b). EditScore-72B has the best overall Winner-set Hit@1 but is the best only on \edittype{Scene}, and Qwen3.8-27B, the most accurate on three groups, is only 5.2 points above random on \edittype{Photometric}. CLIP-I, which prefers the candidate closest to the source, shows why. It is the most accurate judge on \edittype{Photometric} and \edittype{Enhancement}, where the correct result changes little, but it is at or below random on \edittype{Style \& Effects} and \edittype{Geometry}, where the request requires a visible change. What a judge must observe therefore depends on the edit.

Is preference finetuning the remedy? The EditScore~\cite{luo2026editscore} models are trained on large sets of detailed human preference annotations. 
However, off-the-shelf VLMs can surpass EditScore on its own EditReward-Bench~\cite{luo2026editscore} (\cref{tab:vlm-vs-rm}). EditScore models reach 65.9-70.3\%, and with four-pass ensembling~\cite{luo2026editscore}, EditScore-72B reaches 76.3\%. Even judging each candidate alone (\textit{pointwise}), off-the-shelf VLMs can reach 71.7-73.8\%. Furthermore, if we expand the observation space, present candidates side by side in a single judge call (\textit{listwise}), the accuracy can be boosted to 85.2-86.1\%. Such comparative judging is not supported by EditScore models, which are finetuned with specific input and output templates. \textbf{Thus, what a judge sees can matter more than how it is trained}. Here, comparative judging enables expanded observation space, leading to better judgments.

\begin{figure}[t]
    \centering
    \includegraphics[width=\linewidth]{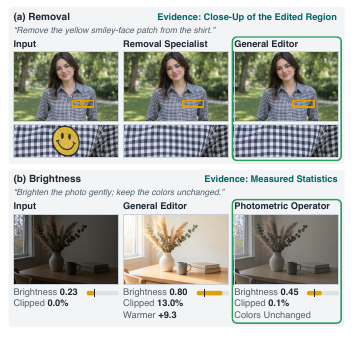}
    \vspace{-2.5em}
    \caption{
        \textbf{Examples of the verification record (evidence).}
        (a) A close-up of the edited region reveals the blurred checks left by the removal specialist.
        (b) Measured brightness, clipping and color shift accurately reveal the performance of the photometric operator.
    }
    \vspace{-1.5em}
    \label{fig:evidence}
\end{figure}

\vspace{-0.5em}
\section{\method}
\vspace{-0.5em}
\label{sec:editagent}

Given a user request, a planner decomposes it into ordered editing steps (\cref{fig:system}). For each step, an analyzer specifies the editing operation and its target, a grounder localizes the target with a VLM and SAM 2.1~\cite{ravi2025sam} when a method needs a mask, and a router selects the applicable editing methods, each of which generates at least one candidate in parallel (\cref{sec:editing_methods}). An off-the-shelf VLM judge evaluates the candidates together with their verification records (\cref{sec:evidence_method}), either accepting one candidate or returning feedback for a retry. Each accepted image becomes the source of the next step, and the requirements it satisfies are carried forward so that subsequent turns can verify that they still hold (\cref{sec:planning}).

\subsection{Expanding the Action Space}

\label{sec:editing_methods}

\noindent \textbf{Candidate Construction.}
Given an executable instruction and its source image, the analyzer specifies the editing operation and recommends candidate methods, and the router turns the recommendation into a set of methods that each contribute at least one candidate. Different methods explore different realizations of the request, while repeated sampling explores variation within a method. 

\noindent \textbf{Action Space.}
Besides the general editor, the capability pool contains: 1) masked local editing and recoloring, which run the general editor inside a SAM 2.1 mask, 2) object removal with OSOR~\cite{zhou2026osor}, 3) object movement with a dedicated model, 4) subject scaling by segmentation and compositing, 5) global and local photometric operators, 6) style filters and neural style transfer, 7) flips, rotations, and rotation to level, and 8) zoom-in by cropping with super-resolution and zoom-out by outpainting. Details can be found in \cref{tab:supp-pool}.

\subsection{Expanding the Observation Space}
\label{sec:evidence_method}

\noindent \textbf{Comparative Judging.}
Inspired by \cref{sec:missed_candidate}, we provide the judge with all candidates in one call. It only needs one VLM call per step, whose comparison yields feedback for the next attempt if a retry is needed.

\noindent \textbf{Verification Records.} Just as a professional retoucher uses different tools to judge an editing operation, we argue that different edits require checks of different properties (\cref{fig:evidence}). For example, a localized edit requires checking both the target change and the preservation of unrelated content, whereas a geometric transformation requires assessing its direction and extent. Some of these properties can be checked by code during execution, some require visual judgment. For each candidate, we collect these checks into a verification record, together with what remains for visual judgment. \method collects the verification record and gives to the judge, so the judge no longer judges by eye alone. Therefore, after all, the judge will receive the edit instruction, the source image, all candidates and their records in one call.

\begin{table}[t]
\centering
\caption{Per-edit-group Winner-Set Hit@1 (\%) of Qwen3.8-27B on our diagnostic benchmark.}
\vspace{-0.5em}
\label{tab:evidence_by_type}
\setlength{\tabcolsep}{4pt}
\resizebox{\linewidth}{!}{
\begin{tabular}{cc ccccccc}
\toprule
Listwise & Evidence & Style & Content & Scene & Photo. & Geom. & Enh. & All \\
\midrule
           &            & 84.4          & 81.7          & 67.5          & 54.2          & 66.4          & 65.0          & 71.6 \\
\checkmark &            & 87.1          & \textbf{87.1} & 72.0          & 47.2          & 78.6          & 36.8          & 73.6 \\
\checkmark & \checkmark & \textbf{87.6} & 86.4          & \textbf{81.9} & \textbf{64.4} & \textbf{81.4} & \textbf{80.3} & \textbf{81.3} \\
\bottomrule
\end{tabular}
}
\vspace{-1em}
\end{table}
\begin{figure}[t]
    \centering
    \includegraphics[width=\linewidth]{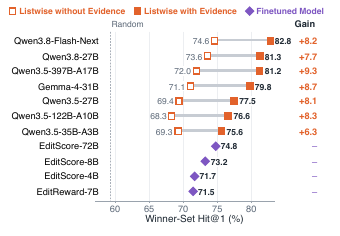}
    \vspace{-2.5em}
    \caption{\textbf{Verification record (evidence) further lifts off-the-shelf VLM judge.} Winner-Set Hit@1 on our diagnostic benchmark.}
    \vspace{-1.5em}
    \label{fig:evaluator-records}
\end{figure}

\noindent \textbf{Performance on the Diagnostic Benchmark.}
To evaluate the effectiveness of the enriched observation space, we report Winner-Set Hit@1 on the 515 candidate sets with at least one usable edit from our diagnostic benchmark (\cref{sec:diagnostic_benchmark}), averaged over three runs.
\cref{tab:evidence_by_type} expands the observation space of the Qwen3.8-27B judge in \cref{fig:motivation}b one step at a time. Listwise comparison raises its accuracy from 71.6\% to 73.6\% and helps on four of the six groups. Comparison alone, however, lowers accuracy on \edittype{Photometric} (from 54.2\% to 47.2\%) and \edittype{Enhancement} (from 65.0\% to 36.8\%), the two groups where the correct result changes little from the source (\cref{sec:missed_candidate}). Adding verification records recovers both groups to 64.4\% and 80.3\%, above the pointwise judge, and raises the overall accuracy to 81.3\%, 6.5 points above EditScore-72B, the strongest judge in \cref{fig:motivation}b. \textbf{Both expansions of the observation space improve the judge, and evidence helps compensate where comparison fails.}

This gain is not specific to Qwen3.8-27B. As shown in \cref{fig:evaluator-records}, seven off-the-shelf VLM judges are improved, by 6.3 to 9.3 points. Without records, none of them reaches the finetuned EditScore-72B (74.8\%). With records, every one of them exceeds all four finetuned reward models, and Qwen3.8-Flash-Next reaches 82.8\%, 8.0 points above EditScore-72B, the strongest judge in \cref{fig:motivation}b.

\begin{table*}[t]
\centering
\vspace{-1em}
\caption{
Comparison of different editing models on EdiVal. Closed-source model numbers from EdiVal. 
Each cell reports turn 1 / turn 2 / turn 3 scores
(0-100).
$^\dagger$Generates at 1024\,px, while all outputs are downsampled to 512$\times$512 for evaluation.
}
\label{tab:edival-main}
\vspace{-0.75em}

\setlength{\tabcolsep}{2em}
\renewcommand{\arraystretch}{1.0}

\resizebox{\linewidth}{!}{%
\begin{tabular}{l c c c}
\toprule
\textbf{Model}
& \textbf{EdiVal-IF} $\uparrow$
& \textbf{EdiVal-CC} $\uparrow$
& \textbf{EdiVal-O} $\uparrow$ \\
\midrule

\multicolumn{4}{l}{\textit{Closed-source}} \\
\quad FLUX.1-Kontext-max$^\dagger$~\cite{bfl2025kontext}
& 69.49 / 46.89 / 31.83
& 93.93 / 90.90 / 88.40
& 80.79 / 65.29 / 53.04 \\
\quad GPT-Image-1$^\dagger$~\cite{openai2025gptimage1}
& 73.12 / 54.89 / 38.35
& 81.00 / 77.78 / 75.50
& 76.96 / 65.34 / 53.81 \\
\quad Nano Banana$^\dagger$~\cite{google2025geminiimage}
& 70.70 / 50.66 / 35.35
& 93.91 / 90.48 / 89.48
& 81.48 / 67.70 / 56.24 \\
\quad FLUX.2-max~\cite{bfl2025flux2max}
& 75.55 / 55.27 / 39.36
& 92.91 / 88.30 / 85.78
& 83.78 / 69.86 / 58.10 \\
\quad Nano Banana 2$^\dagger$~\cite{google2026nanobanana2}
& 73.89 / 54.17 / 38.61
& 93.54 / 90.52 / 88.61
& 83.14 / 70.02 / 58.49 \\
\quad GPT-Image-1.5$^\dagger$~\cite{openai2025gptimage15}
& 75.19 / 55.92 / 40.08
& 94.49 / 91.20 / 88.49
& 84.29 / 71.41 / 59.55 \\
\quad Seedream 4.0$^\dagger$~\cite{seedream2025seedream4}
& 75.93 / 55.58 / 41.59
& 92.51 / 88.03 / 85.86
& 83.81 / 69.95 / 59.76 \\

\midrule

\multicolumn{4}{l}{\textit{Open-source}} \\
\quad AnyEdit~\cite{yu2025anyedit}
& 41.07 / 16.32 / 7.22
& 86.42 / 78.91 / 70.10
& 59.58 / 35.89 / 22.50 \\
\quad UltraEdit~\cite{zhao2024ultraedit}
& 51.37 / 17.70 / 6.36
& 86.80 / 84.50 / 82.40
& 66.78 / 38.67 / 22.89 \\
\quad OmniGen~\cite{xiao2025omnigen}
& 54.72 / 24.48 / 10.66
& 93.00 / 88.42 / 83.92
& 71.34 / 46.52 / 29.91 \\
\quad FLUX.1-Kontext-dev$^\dagger$~\cite{bfl2025kontext}
& 59.97 / 32.69 / 16.61
& 95.32 / 92.24 / 90.22
& 75.61 / 54.91 / 38.71 \\
\quad Step1X-Edit~\cite{liu2025step1x}
& 61.89 / 34.97 / 17.83
& 92.76 / 88.52 / 85.21
& 75.77 / 55.64 / 38.98 \\
\quad Qwen-Image-Edit$^\dagger$~\cite{wu2025qwenimagetechnicalreport}
& 72.90 / 44.06 / 22.55
& 84.22 / 80.52 / 77.98
& 78.36 / 59.56 / 41.93 \\

\addlinespace[2pt]

\quad Qwen-Image-Edit-2511~\cite{wu2025qwenimagetechnicalreport}
& 75.35 / 52.97 / 39.16
& 76.59 / 64.79 / 57.95
& 75.97 / 58.58 / 47.64 \\

\qquad \textbf{+ \method}
& 75.35 / 57.69 / 41.61
& 86.29 / 77.87 / 73.92
& 80.64 / 67.02 / \textbf{55.46} \\

\addlinespace[2pt]

\quad FLUX.2-klein-9B~\cite{bfl2026flux2klein}
& 69.76 / 48.08 / 31.82
& 94.91 / 91.42 / 88.94
& 81.37 / 66.30 / 53.20 \\

\qquad \textbf{+ \method}
& 75.35 / 55.77 / 40.21
& 94.67 / 90.65 / 87.90
& 84.46 / 71.10 / \textbf{59.45} \\

\addlinespace[2pt]

\quad Qwen-Image-Edit-2511$^\dagger$~\cite{wu2025qwenimagetechnicalreport}
& 76.22 / 57.34 / 41.08
& 93.95 / 89.55 / 87.05
& 84.62 / 71.66 / 59.80 \\

\qquad \textbf{+ \method}$^\dagger$
& 76.05 / 57.87 / 44.93
& 93.80 / 89.77 / 87.15
& 84.46 / 72.08 / \textbf{62.58} \\

\bottomrule
\end{tabular}%
}
\vspace{-1em}
\end{table*}

\subsection{Planning and Preserving Edits across Turns}
\label{sec:planning}
\label{sec:multiturn_method}

\begin{figure}[t]
    \centering
    \includegraphics[width=\linewidth]{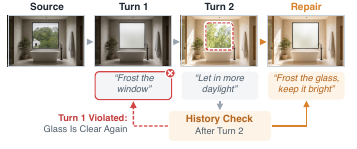}
    \vspace{-2em}
    \caption{
        \textbf{The history check extends what the judge observes to earlier turns.}
        Letting in more daylight in turn 2 clears the frosted glass from turn 1. The history check detects this and repairs.
    }
    \vspace{-1em}
    \label{fig:multistep_multiturn}
\end{figure}

The planner splits a request into dependent steps when its parts need separate execution, and each accepted step becomes the source of the next. Because the step judge sees one step at a time, a later edit can undo an earlier one unnoticed. After every turn, a history check therefore shows the VLM the original image, the results of earlier turns and the current image, and asks whether each earlier request still holds. On a confident violation, \method runs one repair turn and adopts it only if its step judge accepts it (\cref{fig:multistep_multiturn}).

\section{Experiments}
\label{sec:experiments}

\subsection{Experimental Setup}
\label{sec:exp-setup}

\noindent \textbf{Benchmarks.}
We evaluate \method on three benchmarks covering multi-turn and complex-instruction editing. EdiVal~\cite{chen2025edival} comprises 572 three-turn sessions and assesses editing quality as successive requests accumulate. Complex-Edit~\cite{yang2025complexedit} comprises 8,496 real and synthetic examples spanning different levels of instruction complexity. ComplexBench-Edit~\cite{wang2025complexbenchedit} comprises 763 examples with compositional instructions, including parallel constraints and multi-step chains. Together, they assess performance across editing turns, complexity, and compositional structure.

\noindent \textbf{Backbones and Baselines.}
We instantiate \method with two general-purpose editors, Qwen-Image-Edit-2511~\cite{wu2025qwenimagetechnicalreport} and FLUX.2-klein-9B~\cite{bfl2026flux2klein}. For each backbone, we compare standalone editing with its integration into \method. We also compare it with seven agentic editors (\cref{tab:edival-main-agent}). On EdiVal, both backbones edit at the 512\,px input resolution. Because most closed-source editors generate at 1024\,px or more, we also run Qwen-Image-Edit-2511 at 1024\,px and resize back to 512\,px for fair comparison. The comparison with agentic editors and all ablations use the 512\,px setting.

\noindent \textbf{Evaluation.}
We follow the evaluation rubrics of each benchmark. EdiVal uses its released judges, and Qwen3.8-27B~\cite{qwen38} replaces GPT-4o and Gemini 2.0 Flash Thinking on Complex-Edit and ComplexBench-Edit. As it is also our internal judge, to avoid potential bias, we additionally evaluate \method using Gemma-4-31B~\cite{team2026gemma}  (\cref{tab:supp-gemma}).

\noindent \textbf{Implementation Details.}
Qwen3.8-27B serves as the planner, analyzer, grounder and judge, and SAM 2.1~\cite{ravi2025sam} turns predicted boxes into masks. Each attempt generates up to three candidates, at least one per applicable method, shown to the judge in random order. Each step allows two retries and each turn two plan revisions.

\subsection{\method as a General Editing Framework}
\label{sec:exp-framework}

\noindent \textbf{Multi-turn Editing.}
Without finetuning, \method improves EdiVal-O (\cref{tab:edival-main}) across both baselines. Specifically, Qwen-Image-Edit-2511 at the 1024 px achieves the best performance, increasing from 59.80 to 62.58 at Turn 3.

\begin{table*}[t]
\centering
\caption{Editing performance on ComplexBench-Edit (scores 0-100) and Complex-Edit (scores 0-10).}
\label{tab:complex-main}
\vspace{-0.75em}
\fontsize{8.5}{10.2}\selectfont
\setlength{\tabcolsep}{4pt}
\begin{tabular*}{\linewidth}{@{\extracolsep{\fill}} l cccccc cccc @{}}
\toprule
\multirow{2}{*}{\textbf{Model}} & \multicolumn{6}{c}{\textbf{ComplexBench-Edit}} & \multicolumn{4}{c}{\textbf{Complex-Edit}} \\
\cmidrule(lr){2-7}\cmidrule(lr){8-11}
 & Obj. $\uparrow$ & Obj.-At. $\uparrow$ & Obj.-At.-G. $\uparrow$ & Two-chain $\uparrow$ & Three-chain $\uparrow$ & Avg.\,$\uparrow$ & IF $\uparrow$ & IP $\uparrow$ & PQ $\uparrow$ & O\,$\uparrow$ \\
\midrule
Qwen-Image-Edit-2511 & 56.98 & 60.81 & 67.97 & 57.59 & 28.67 & 54.40 & 8.721 & 6.259 & 7.132 & 7.371 \\
\quad \textbf{+ EditAgent}    & 68.92 & 70.74 & 73.17 & 60.21 & 31.11 & \textbf{60.83} & 8.736 & 6.391 & 7.209 & \textbf{7.445} \\
\midrule
FLUX.2-klein-9B      & 67.52 & 70.56 & 73.87 & 61.67 & 33.61 & 61.45 & 8.763 & 7.304 & 6.964 & 7.677 \\
\quad \textbf{+ EditAgent}    & 70.05 & 74.69 & 76.80 & 64.48 & 36.53 & \textbf{64.51} & 8.783 & 7.131 & 7.224 & \textbf{7.713} \\
\bottomrule
\end{tabular*}
\vspace{-0.5em}
\end{table*}

\begin{table*}[t]
\centering
\caption{
Comparison with agentic editors on EdiVal. We report EdiVal-O at turns 1/2/3 and inference cost (H100 GPU hour).
All agents use the same base editors and Qwen3.8-27B~\cite{qwen38} as the VLM.
Best \textbf{bold}, second \underline{underlined}.
$^\dagger$Re-implemented from the paper.
}
\label{tab:edival-main-agent}
\vspace{-0.75em}
\fontsize{8.5}{10.2}\selectfont
\setlength{\tabcolsep}{4pt}
\renewcommand{\arraystretch}{1.0}
\begin{tabular*}{\linewidth}{@{\extracolsep{\fill}} l cccc cccc @{}}
\toprule
\multirow{2}{*}{\textbf{Model}}
& \multicolumn{4}{c}{\textbf{Qwen-Image-Edit-2511}~\cite{wu2025qwenimagetechnicalreport}}
& \multicolumn{4}{c}{\textbf{FLUX.2-klein-9B}~\cite{bfl2026flux2klein}} \\
\cmidrule(lr){2-5}\cmidrule(lr){6-9}
& Turn 1 $\uparrow$ & Turn 2$\uparrow$ & Turn 3 $\uparrow$ & Inference Cost $\downarrow$
& Turn 1 $\uparrow$ & Turn 2$\uparrow$ & Turn 3 $\uparrow$ & Inference Cost $\downarrow$ \\
\midrule
Standalone Editor & 75.97 & 58.58 & 47.64 & 1.1 & 81.37 & 66.30 & 53.20 & 0.5 \\
\midrule
MIRA$^\dagger$~\cite{zeng2026mira}
& 75.79 & 57.59 & 45.59 & 11.3 & 82.57 & 67.08 & 53.76 & 5.6 \\
IMAGAgent~\cite{shen2026imagagent}
& 76.83 & 60.15 & 49.78 & 10.9 & 82.63 & 68.00 & 55.21 & 4.6 \\
EditThinker~\cite{li2025editthinker}
& 76.58 & 60.47 & 49.23 & 9.4 & 82.62 & 67.20 & 55.97 & 4.6 \\
GenArtist$^\dagger$~\cite{wang2024genartist}
& \underline{79.51} & 62.94 & 50.54 & 9.5 & 82.94 & 67.48 & 54.74 & 5.2 \\
VisionDirector-TF$^\dagger$~\cite{chu2026visiondirector}
& 76.68 & 59.58 & 48.68 & 9.9 & 82.19 & 68.77 & 56.96 & 4.8 \\
RISE-Agent~\cite{yang2026riseagent}
& 78.86 & \underline{63.66} & \underline{52.10} & 14.7 & \underline{83.95} & \underline{70.30} & \underline{58.42} & 7.2 \\
\midrule
\textbf{\method}
& \textbf{80.64} & \textbf{67.02} & \textbf{55.46} & 12.9
& \textbf{84.46} & \textbf{71.10} & \textbf{59.45} & 6.3 \\
\bottomrule
\end{tabular*}
\vspace{-1em}
\end{table*}

\begin{figure*}[t]
  \centering
  \vspace{-0.5em}
  \includegraphics[width=\textwidth]{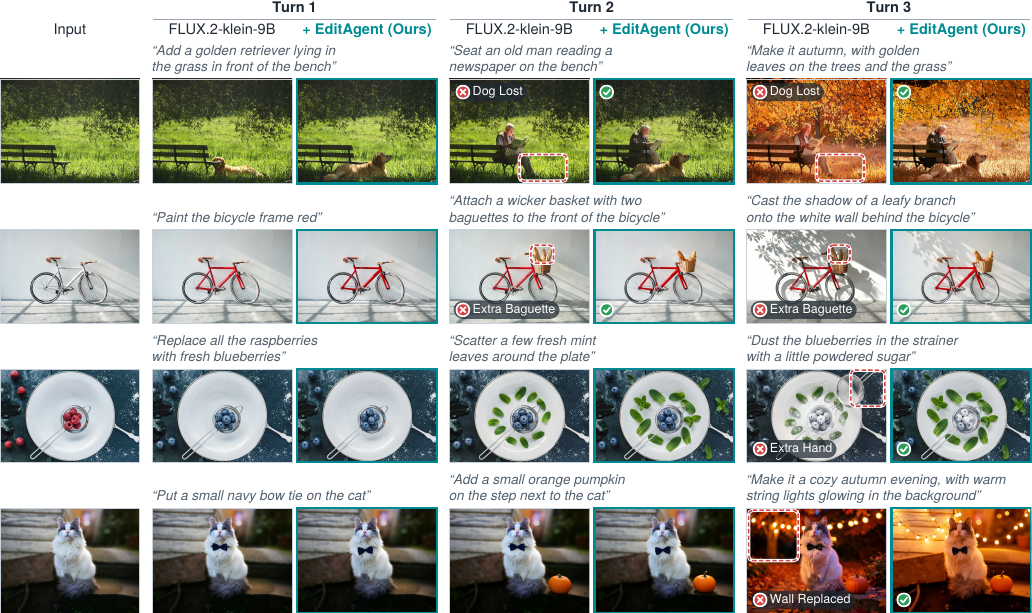}
  \vspace{-1.75em}
  \caption{Multi-turn editing comparison between FLUX.2-klein-9B and \method.}
  \vspace{-0.75em}
  \label{fig:qual-multi}
\end{figure*}
\begin{table*}[t]
\centering
\caption{Training data from EditAgent versus data generated directly by FLUX.2-klein-9B. Each baseline model is fully finetuned for one epoch; ``---'' denotes no finetuning. To bridge the train-eval gap of prompting, we rewrite the input prompts for EdiVal.}
\label{tab:data-engine}
\vspace{-0.5em}

\setlength{\tabcolsep}{1.5em}
\renewcommand{\arraystretch}{1.0}

\resizebox{\linewidth}{!}{%
\begin{tabular}{llccc}
\toprule
\textbf{Baseline} & \textbf{Training Data Source} & \textbf{EdiVal-O Turn 3} $\uparrow$ & \textbf{Complex-Edit O} $\uparrow$ & \textbf{ComplexBench-Edit Avg} $\uparrow$ \\
\midrule
\multirow{3}{*}{Qwen-Image-Edit-2511}
& --- & 49.66 & 7.371 & 54.40 \\
& FLUX.2-klein-9B & 48.51 & 7.334 & 55.84 \\
& \textbf{EditAgent} & \textbf{51.61} & \textbf{7.435} & \textbf{57.57} \\
\midrule
\multirow{3}{*}{FLUX.2-klein-base-9B}
& --- & 54.53 & 7.517 & 57.84 \\
& FLUX.2-klein-9B & 52.83 & 7.351 & 57.47 \\
& \textbf{EditAgent} & \textbf{56.53} & \textbf{7.615} & \textbf{59.98} \\
\bottomrule
\end{tabular}%
}
\vspace{-1.25em}
\end{table*}

\noindent \textbf{Complex-instruction Editing.}
On ComplexBench-Edit (\cref{tab:complex-main}), the average rises by 6.43 and 3.06 points for Qwen-Image-Edit-2511 and FLUX.2-klein-9B, respectively, improving all five groups. The Complex-Edit overall score rises by 0.074 and 0.036. Across benchmarks, specialized methods deliver 19\% to 31\% of the accepted steps. 

\noindent \textbf{Comparison with Agentic Editors.} In \cref{tab:edival-main-agent}, with the same backbones and VLM, \method achieves the best turn 3 score (59.45, against 58.42 for the strongest agent, RISE-Agent) and the best score in all the 3 turns.

\subsection{\method as a Data Engine}
\label{sec:exp-data}

Because \method flags every edit that its judge did not accept, its fully accepted sessions can serve as training data. Using FLUX.2-klein-9B as the general editor, \method produces about 25.7K sessions, averaging four turns each. Each session yields one training sample consisting of the source image, a session-level instruction, and the final edited image. For comparison, we use FLUX.2-klein-9B to directly edit the same sessions, keeping the training inputs identical and varying only the final images. We fully finetune Qwen-Image-Edit-2511 and FLUX.2-klein-base-9B on each dataset for one epoch and evaluate them as standalone editors.

As shown in \cref{tab:data-engine}, training on \method data improves both editors across all three benchmarks. On EdiVal-O at turn 3, Qwen-Image-Edit-2511 and FLUX.2-klein-base-9B gain 1.95 and 2.00 points, respectively, while their ComplexBench-Edit averages increase by 3.17 and 2.14 points. More importantly, \method data consistently outperforms directly generated FLUX.2-klein-9B data under the same training recipe. 
Notably, direct FLUX.2-klein-9B data degrades FLUX.2-klein-base-9B on all three benchmarks, whereas \method data improves it. 
These results suggest that \method provides higher-quality training supervision rather than merely additional training examples.

\begin{table}[h]
\centering
\caption{Component ablation on EdiVal with Qwen-Image-Edit-2511. Rows with $K{=}n$ replace the action space with $n$ identical copies of the general editor. The last row shows the full \method.}
\label{tab:component-ablation}
\vspace{-0.5em}
\setlength{\tabcolsep}{0.6em}
\renewcommand{\arraystretch}{1}
\resizebox{\linewidth}{!}{%
\begin{tabular}{ccccc}
\toprule
\multirow[c]{2}{*}{\shortstack[c]{\textbf{Vanilla \&}\\\textbf{Agent}}}
& \multirow[c]{2}{*}{\shortstack[c]{\textbf{Action}\\\textbf{Space}}}
& \multicolumn{2}{c}{\textbf{Observation Space}}
& \multirow[c]{2}{*}{\textbf{EdiVal-O (T3)}} \\
\cmidrule(lr){3-4}
& &
\shortstack[c]{\textbf{Parallel}\\\textbf{Comparison}}
& \shortstack[c]{\textbf{Verification}\\\textbf{Record}}
& \\
\midrule

$\times$ & $\times$ & $\times$ & $\times$ & 47.64 \\
$\checkmark$ & $\times$ & $\times$ & $\times$ & 49.43 \\
$\checkmark$ & $\checkmark$ & $\times$ & $\times$ & 52.37 \\
$\checkmark$ & $\checkmark$ & $\checkmark$ & $\times$ & 53.18 \\

\midrule
$\checkmark$ & $K{=}2$ & $\checkmark$ & $\times$ & 47.53 \\
$\checkmark$ & $K{=}4$ & $\checkmark$ & $\times$ & 48.90 \\
$\checkmark$ & $K{=}8$ & $\checkmark$ & $\times$ & 49.01 \\

\midrule
$\checkmark$ & $\checkmark$ & $\checkmark$ & $\checkmark$
& \textbf{55.46} \\

\bottomrule
\end{tabular}%
}
\vspace{-2em}
\end{table}

\subsection{Ablation Study}
\label{sec:exp-ablation}

\cref{tab:component-ablation} evaluates each component of \method. The vanilla agent, equipped with planning, analysis, history checking, and repair/retry, improves EdiVal-O at turn 3 from 47.64 to 49.43. Expanding the action space yields a 2.94-point gain. Sampling the general editor more often does not reproduce this gain. If we replace the action space with $K$ copies of the general editor and keep parallel comparison, EdiVal-O reaches 47.53, 48.90, and 49.01 for $K=2$, $4$, and $8$. This agrees with the diagnostic study in \cref{sec:missing_candidate}, where restricting \method to the general editor lowers the share of usable delivered edits. Expanding the observation space further improves performance by 3.09 points.

\subsection{Qualitative Results}
\label{sec:qualitative-results}

\Cref{fig:qual-multi} compares standalone FLUX.2-klein-9B with \method. For multi-turn editing, each turn edits the previous output, so an error carries into every later turn. Standalone editing drops the dog of turn 1 when it adds the old man in turn 2. When turn 2 asks for two baguettes, it adds three, which remains in turn 3. Instead, \method judges each step against its source image, which already contains the earlier edits, and checks that they remain. More qualitative comparisons is provided in \cref{sec:supp-qualitative}, further demonstrating the overall advantages of our method, even over GPT-Image-2.
\vspace{-0.5em}
\section{Conclusion}
\vspace{-0.5em}
\label{sec:conclusion}

We presented \method, which rethinks agentic image editing as expanding two spaces: the action space of the agent and the observation space of its judge. It routes each step to general-purpose and specialized editing methods, generates their candidates in parallel, and lets the off-the-shelf VLM judge compare them with verification records. \method improves two editing backbones across three benchmarks, and functions as a robust data engine, leveraging its accepted sessions to improve standalone editors through fine-tuning.

\noindent \textbf{Limitations.} 
\method incurs roughly 10$\times$ the inference time of standalone editing models, and its performance is inherently bounded by the capabilities of its underlying editing methods. Nevertheless, \method can serve as a data engine for finetuning stronger and more efficient standalone editors, potentially amortizing its computational cost.

{
    \small
    \bibliographystyle{ieeenat_fullname}
    \bibliography{main,supp}
}

\clearpage
\clearpage
\raggedbottom
\setcounter{page}{1}
\setcounter{section}{0}
\renewcommand{\thesection}{\Alph{section}}
\renewcommand{\theHsection}{supp.\Alph{section}}
\setcounter{figure}{0}\renewcommand{\thefigure}{S\arabic{figure}}\renewcommand{\theHfigure}{supp.\arabic{figure}}
\setcounter{table}{0}\renewcommand{\thetable}{S\arabic{table}}\renewcommand{\theHtable}{supp.\arabic{table}}

\newcommand{\supppart}[1]{\multicolumn{4}{@{}l}{\rule{0pt}{3.2ex}\large\textbf{#1}} \\[3pt]}
\newcommand{\suppentry}[4]{%
  \textbf{\ref*{#1}} & \hyperref[#1]{\textbf{#2}}\par\vspace{1pt}{\small\color{gray!25!black}#4} & #3 & \pageref*{#1} \\[5pt]}

\twocolumn[{%
\centering
\Large\textbf{\thetitle}\\
\vspace{0.5em}Supplementary Material\\
\vspace{2.5em}
\normalsize\raggedright
{\Large\textbf{Contents}}\par\vspace{0.6em}
\begin{tabularx}{\textwidth}{@{}p{1.4em} X p{9.5em} r@{}}
 & & \textbf{Main Paper} & \textbf{Page} \\
\midrule
\supppart{Diagnostic Studies}
\suppentry{sec:supp-diagnostic}{Diagnostic Benchmark}{Secs.~\ref*{sec:diagnostic_benchmark} and \ref*{sec:missing_candidate}}{Request design, candidate generation, annotation by professional editors, the Winner-Set Hit@1 metric and the 30 edit types (\cref{tab:supp-edit-types}).}
\suppentry{sec:supp-search}{Parallel Search versus Sequential Refinement}{Sec.~\ref*{sec:search}}{Requests, the two search policies, the human study and the cost of each policy.}
\suppentry{sec:supp-evaluator}{Judge Studies}{Secs.~\ref*{sec:missed_candidate} and \ref*{sec:evidence_method}}{Protocol, records, models, metrics, what records contribute and EditReward-Bench.}
\supppart{Method Details}
\suppentry{sec:supp-workflow}{Workflow Details}{Secs.~\ref*{sec:editagent} and \ref*{sec:planning}}{Roles of the modules (\cref{tab:supp-roles}), the turn and step algorithms (\cref{alg:supp-turn,alg:supp-step}), acceptance, robustness and the default settings (\cref{tab:supp-settings}).}
\suppentry{sec:supp-routing}{Routing and Capability Pool}{Sec.~\ref*{sec:editing_methods}}{Routing rules (\cref{tab:supp-routing}) and the editing methods of the capability pool (\cref{tab:supp-pool}).}
\suppentry{sec:supp-records}{Verification Records}{Sec.~\ref*{sec:evidence_method}}{Close-ups, measurements and code checks (\cref{tab:supp-measurements}).}
\supppart{Experiments}
\suppentry{sec:supp-benchmarks}{Benchmark Experiments}{Secs.~\ref*{sec:experiments}.1 and \ref*{sec:experiments}.2}{Backbones, inputs, scoring, agentic baselines (\cref{tab:supp-agents}), an alternative internal judge (\cref{tab:supp-gemma}), and route usage (\cref{tab:route-usage-supp}).}
\suppentry{sec:supp-data}{Data Engine}{Sec.~\ref*{sec:experiments}.3}{Sources and requests, generation and filtering with pass rates (\cref{tab:supp-data-stats}), compute, training pairs, baseline data and finetuning.}
\suppentry{sec:supp-qualitative}{Additional Qualitative Results}{Sec.~\ref*{sec:qualitative-results}}{More multi-turn and single-turn comparisons (\crefrange{fig:supp-qual-multi-qie}{fig:supp-qual-single-flux}).}
\supppart{Prompts}
\suppentry{sec:supp-prompts}{Prompts}{Sec.~\ref*{sec:editagent}}{Prompts of the planner, analyzer, grounding, judge, re-planner and history check of \method (\crefrange{lst:planner-system}{lst:history-system}), with the names they use for terms of the paper (\cref{tab:supp-glossary}).}
\bottomrule
\end{tabularx}\par
\vspace{0.8em}
{\small Lettered sections, S-numbered figures and tables, algorithms and listings belong to this supplement. Numbered sections, figures and tables refer to the main paper. The sections follow the order of the main paper.\par}
}]
\clearpage

\section{Diagnostic Benchmark}
\label{sec:supp-diagnostic}

\noindent \textbf{Requests.}
The benchmark covers 30 edit types in six groups, with 20 requests per type (\cref{tab:supp-edit-types}). Source images are drawn from COCO val2017~\cite{lin2014coco}, and each image is used in one set only. A VLM (Qwen3.8-27B) screens each photograph so that every edit type is assigned only to photographs that support it, and then writes one natural instruction of the assigned type, following the type's definition in \cref{tab:supp-edit-types}. Requests are written before any candidate exists.

\noindent \textbf{Candidates.}
Candidates come from two runs of \method on each request, one with the full capability pool and one restricted to the general editor. Both runs use FLUX.2-klein-9B as the general editor and Qwen3.8-27B in all VLM roles. All distinct images generated in these runs form the pool of a set, from which we keep up to four, giving priority to the delivered results and to methods not yet represented. This yields 2 to 4 candidates per set and 1,995 candidates in total.

\noindent \textbf{Annotation.}
Five professional Photoshop editors label each set together in a shared interface and agree on one label, so each set has one consensus label. The interface shows the source image, the instruction and the candidates under neutral keys (C1 to C4), in an order fixed by a hash and without any indication of their method. For each set, the editors select every usable candidate, \textit{i.e.}, every candidate that completes the instruction acceptably. A usable candidate is selected even when another candidate is better, so an unselected candidate is unusable, not merely second best. The editors can also mark the set as all usable or all unusable. Of the 600 sets, 248 have one selected candidate, 155 have several, 116 are marked as all usable and 81 as all unusable.

\noindent \textbf{Winner-Set Hit@1.}
The winner set of a set contains its selected candidates, or all candidates when the set is marked all usable. It is thus the set of usable candidates, and Winner-Set Hit@1 measures how often a judge's top choice is usable. We exclude the 81 sets without a usable candidate. Of the remaining 519, we also exclude 4 in which one candidate has no verification record, so that every judge is compared on the same 515 sets. A judge that scores candidates separately selects its top-scoring candidate. When several candidates share the top score, the tie is broken uniformly at random, and we count the expected outcome, \textit{i.e.}, the fraction of the tied candidates that are in the winner set. An invalid output, or a missing score for any candidate of the set, counts as 0. Accuracy is computed per run over the fixed set of sets and averaged over three runs. The random baseline is the mean over sets of the winner-set size divided by the number of candidates. It is 59.32\% on the 515 sets and 47.10\% on the 396 sets with at least one unusable candidate.

\begin{figure}[t]
    \centering
    \includegraphics[width=\columnwidth]{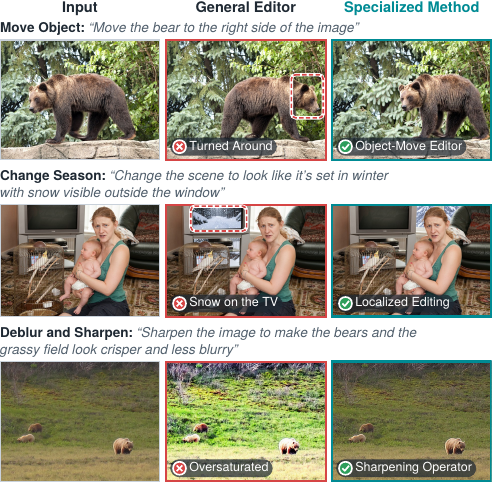}
    \vspace{-1.5em}
    \caption{Diagnostic sets in which only the specialized candidate is usable~(\checkmark). The general editor turns the bear around, puts the snow on the TV screen, and changes the saturation.}
    \vspace{-1.5em}
    \label{fig:generation_failures}
\end{figure}

\noindent \textbf{Specialized Winners.}
\Cref{fig:generation_failures} shows sets in which only the specialized candidate is usable.

\begin{table*}[t]
\centering
\caption{The 30 edit types of the diagnostic benchmark and the definitions given to the instruction writer. Each type has 20 sets. General counts the sets with at least one usable general-editor candidate from either run, and Hit@1 counts the sets used for Winner-Set Hit@1.}
\label{tab:supp-edit-types}
\vspace{-0.5em}
\footnotesize
\setlength{\tabcolsep}{2pt}
\begin{tabularx}{\textwidth}{@{}l l Y r r@{}}
\toprule
\textbf{Group} & \textbf{Edit Type} & \textbf{Definition} & \textbf{General} & \textbf{Hit@1} \\
\midrule
\multirow{9}{*}{Content} 
& Add Text & add readable text/writing/signage somewhere it fits naturally & 19 & 19 \\
 & Remove Text & remove existing text/logo/signage (only if some is visible) & 15 & 15 \\
 & Replace Text & replace visible text with different text (only if text is visible) & 14 & 15 \\
 & Change Text Style & change color/font/size/style of visible text & 9 & 12 \\

 & Add Object & add a new object/animal/person that fits the scene & 18 & 18 \\
 & Remove Object & remove an existing object/person/animal & 15 & 17 \\
 & Replace Object & swap one object for a different one & 16 & 16 \\
 & Move Object & move an existing object/subject to a different position in the same scene & 6 & 11 \\
 & Change Object Appearance & change an object's color/material/texture/state (not swap it) & 17 & 19 \\
\midrule
\multirow{4}{*}{Style \& Effects} & Create Line Art & turn the photo into a clean line/edge drawing or pencil sketch of itself & 20 & 20 \\
 & Apply Style & apply a named artistic style or photographic look to the whole image & 20 & 20 \\
 & Apply Blur & blur the background or the whole image (e.g. bokeh, motion blur) & 17 & 17 \\
 & Convert Color Mode & colorize (if grayscale) or convert to black \& white / grayscale & 16 & 18 \\
\midrule
\multirow{5}{*}{Scene} & Replace Background & replace the background/scene behind the subject & 18 & 18 \\
 & Change Time of Day & change the time of day (e.g. day to night, sunset) & 16 & 16 \\
 & Change Background Appearance & change an attribute of the background only & 15 & 15 \\
 & Change Season & change the season (e.g. summer to winter) & 13 & 14 \\
 & Change Color Tone & shift the overall color tone (warmer/cooler/teal-orange/color grade) & 2 & 18 \\
\midrule
\multirow{4}{*}{Photometric} & Change Lighting & change the lighting (e.g. golden hour glow, harsh noon, soft studio) & 13 & 17 \\
 & Adjust Saturation & increase or decrease color saturation & 11 & 20 \\
 & Adjust Brightness & brighten or darken the whole image noticeably & 9 & 16 \\
 & Adjust Contrast & increase or decrease global contrast & 7 & 19 \\
\midrule
\multirow{6}{*}{Geometry} & Change Viewpoint & change the camera viewpoint/angle (e.g. eye-level to low angle) & 11 & 14 \\
 & Zoom In & zoom in / crop closer onto a named target & 11 & 18 \\
 & Zoom Out & zoom out / widen the view, generating new surroundings (outpaint) & 5 & 16 \\
 & Flip Horizontally & mirror/flip the image horizontally & 1 & 20 \\
 & Extend Canvas & extend the canvas beyond the current frame on one or more sides & 0 & 18 \\
 & Rotate Image & rotate the image (e.g. 90 degrees or a slight straighten) & 0 & 20 \\
\midrule
\multirow{2}{*}{Enhancement} & Denoise and Restore & restore/clean the photo (denoise, remove scratches, fix exposure) & 8 & 19 \\
 & Deblur and Sharpen & sharpen / deblur the image & 2 & 20 \\
\midrule
\multicolumn{3}{@{}l}{Total} & 344 & 515 \\
\bottomrule
\end{tabularx}
\vspace{-1em}
\end{table*}

\section{Refine Sequentially or Search in Parallel}
\label{sec:supp-search}

\noindent \textbf{Requests.}
We sample 48 requests at each of the complexity levels 1, 4 and 8 from the real-image split of Complex-Edit~\cite{yang2025complexedit}, each with its own source image. A request at level $k$ chains $k$ atomic instructions, so the three levels range from a single edit to a long compound request.

\noindent \textbf{Policies.}
Both policies use FLUX.2-klein-9B, Qwen3.8-27B as the VLM, and at most four images per request. Sequential refinement generates the first image from the source with the original instruction. After each of the first three images, it asks the VLM for a corrective request for the latest image and applies it to that image, and a final selection call picks among the four images (3 feedback calls and 1 selection call). Parallel search generates two images from the source with the original instruction. If the VLM accepts neither, it revises the instruction and generates two more images from the source before a final selection, using at most two VLM calls. The VLM sees the source, the candidate images and their difference maps, in a shared random order. Sequential refinement re-edits the latest image, as iterative editing agents commonly do, whereas the retries of \method re-edit the source of the step with a revised instruction (\cref{sec:supp-workflow}).

\noindent \textbf{Human Study.}
Five professional Photoshop editors view the final images of the policies blindly, next to the source and the request, in random positions and without labels, and assess instruction following, preservation of unrelated content, and artifacts. For each comparison of parallel search with sequential refinement, the editors either choose the better policy or mark the two as equal. Parallel search is preferred in 102 comparisons and sequential refinement in 28, with 14 ties, so parallel search wins 78.5\% of the 130 comparisons in which the editors prefer one policy.

\noindent \textbf{Cost.}
With at most four images per request for both policies, parallel search needs at most two VLM calls per request, and sequential refinement always needs four.

\section{Observation Space}
\label{sec:supp-evaluator}

\noindent \textbf{Protocol.}
The judges of \cref*{sec:missed_candidate} are the reward models with their released inference and Qwen3.8-27B with the pointwise judge, ranked by the probability of its pass verdict. They score each candidate separately, and tied top-scoring candidates share the credit. The listwise judge uses the instructions of \cref{lst:judge-system} and receives the request, the source image and all candidates of a set in one call. The pointwise judge uses the same instructions for a single candidate. The candidate order of each set is fixed across conditions and models and shifts cyclically by one position per run.

\noindent \textbf{Records in These Studies.}
The studies use the verification records of \cref{sec:supp-records} (\cref{tab:supp-measurements}), computed for the labeled candidates of each set. For a localized request, the region that the request names is localized once in the source image, and every candidate gets a close-up of this region.

\noindent \textbf{Models and Inference.}
\Cref*{fig:evaluator-records} lists the VLMs and reward models. Each VLM is served with vLLM~\cite{kwon2023vllm} from its released checkpoint and judges every set three times, with thinking enabled and the sampling settings of its generation configuration. A call without a valid answer counts as a miss. EditScore uses its released inference over three runs. EditReward-7B uses its released deterministic scorer in one run. In the benchmark experiments, every VLM role uses the settings of \cref{tab:supp-settings}.

\noindent \textbf{Metrics.}
CLIP-I~\cite{radford2021clip}, used in \cref*{sec:missed_candidate} of the main paper, is the cosine similarity between the CLIP ViT-L/14 image embeddings of the source and the candidate.

\noindent \textbf{What Records Contribute.}
Records improve each of the seven VLMs of \cref*{fig:evaluator-records}, by 6.3 to 9.3 points, and raise their mean Winner-Set Hit@1 from 71.17\% to 79.25\%. On the 396 sets with at least one unusable candidate, the mean rises from 62.60\% to 73.06\%, against 47.10\% for a random choice. The gain is largest on \edittype{Enhancement}, \edittype{Photometric} and \edittype{Scene} edits (40.7, 20.5 and 9.5 points on average).

\noindent \textbf{EditReward-Bench.}
We use the EditReward-Bench of EditScore~\cite{luo2026editscore}, which is distinct from the EditReward model~\cite{wu2026editreward}. It has 3,072 human-labeled candidate pairs over 13 task types, with preferences for prompt following, consistency and overall quality, of which \cref*{tab:vlm-vs-rm} reports overall quality. A pair counts as correct when the human-preferred candidate ranks higher, and a tie counts as wrong. The pointwise column of \cref*{tab:vlm-vs-rm} uses the official prompt of the benchmark, which gives each candidate an integer score. The listwise judge shows both candidates in one call and uses the rubric and output format of \cref{lst:judge-system}, without records. Every VLM runs three times. 

\section{Agentic Workflow Details}
\label{sec:supp-workflow}

This section details the control flow of \method summarized in \cref{sec:editagent,sec:planning} of the main paper. It gives the roles and what each receives (\cref{tab:supp-roles}), the turn and step loops (\cref{alg:supp-turn,alg:supp-step}) and the default settings (\cref{tab:supp-settings}).

\begin{table*}[t]
\centering
\caption{Roles in \method. In the benchmark experiments and in the runs that produced the diagnostic candidates, every VLM role uses Qwen3.8-27B with the settings of \cref{tab:supp-settings}. The data engine uses Qwen3.8-27B as the judge (\cref{sec:supp-data}).}
\label{tab:supp-roles}
\vspace{-0.5em}
\small
\begin{tabularx}{\textwidth}{@{}l l Y Y l@{}}
\toprule
\textbf{Role} & \textbf{Model} & \textbf{Input} & \textbf{Output} & \textbf{Prompt} \\
\midrule
Planner & VLM & Request of the turn, the version graph and the carried requirements & Goal and ordered steps, each with its operation, target, regions to keep, parameters and new requirements & \cref{lst:planner-system} \\
Analyzer & VLM & One step, source image of the step, carried requirements, feedback on earlier attempts, reports of earlier steps, parameter documentation of the applicable methods & Explicit instruction stating what changes and what stays, operation, target, regions to keep, parameters, acceptance criteria, candidate count & \cref{lst:analyzer-system} \\
Router & Code & Analyzed step & Methods, each contributing at least one candidate (\cref{tab:supp-routing}) & None \\
Grounder & VLM and SAM 2.1 & Image and a referring expression & One box per referred object in coordinates normalized to $[0,1000]$, refined into a mask by SAM 2.1 and reviewed by the VLM & \cref{lst:ground,lst:mask-review} \\
Editing Methods & \cref{tab:supp-pool} & Source image of the step, instruction, target, parameters & Candidate image and its execution artifacts, such as a mask & None \\
Judge & VLM & Request of the step with its scope, acceptance criteria and regions to keep, source image, and all candidates with their verification records in a shuffled order & A note, a pass/fail verdict and a score from 1 to 10 for every candidate, then the best candidate and a ranking & \cref{lst:judge-system} \\
Re-Planner & VLM & Request, finished steps with reports, failed step with the feedback on all its attempts, pending steps & New remaining steps, or none to keep the best attempt & \cref{lst:replanner-system} \\
History Check & VLM & Original image, results of earlier turns, current image, then-and-now close-ups, requirements established in earlier turns & For each earlier turn a verdict (still holds or superseded), score, confidence and evidence, and a repair instruction when needed & \cref{lst:history-system} \\
\bottomrule
\end{tabularx}
\vspace{-1em}
\end{table*}

\begin{table}[t]
\centering
\caption{Default settings of \method, used in all benchmark experiments and in the data engine (EdiVal images at $512^2$ px).}
\label{tab:supp-settings}
\vspace{-0.5em}
\small
\setlength{\tabcolsep}{4pt}
\begin{tabularx}{\columnwidth}{@{}Y r@{}}
\toprule
\textbf{Setting} & \textbf{Value} \\
\midrule
Candidates per Attempt & at most 3 \\
Attempts per Step & 3 \\
Re-Plans per Turn & 2 \\
Repairs per Turn & 1 \\
Repair Confidence & at least 0.4 \\
Seed of Candidate $j$ in Attempt $a$ & $32 + 7a + j$ \\
\midrule
Working Size, Multiple of 32 px & at most $1024^2$ px \\
Earlier Results Shown to History Check & at most 5 \\
Close-Ups for History Check & at most 4 \\
Mask Dilation for Masked Edit & 24 px \\
Feather for Masked Edit & 6 px \\
Mask Dilation for Masked Recolor & 8 px \\
Feather for Masked Recolor & 4 px \\
\midrule
VLM Temperature & 0.2 \\
VLM Output Tokens & at most 4,096 \\
VLM Context Length & 16,384 tokens \\
Images per VLM Request & at most 20 \\
Thinking & disabled \\
Output Format & JSON \\
Grounding Coordinates & $[0, 1000]$ \\
Segmentation & SAM 2.1 \\
Mask Reviews per Target & at most 2 \\
\bottomrule
\end{tabularx}
\vspace{-1em}
\end{table}

\begin{table}[t]
\centering
\caption{Methods raced for each kind of step, with the operation that the planner and analyzer assign to it.}
\label{tab:supp-routing}
\vspace{-0.5em}
\footnotesize
\setlength{\tabcolsep}{3pt}
\begin{tabularx}{\columnwidth}{@{}>{\hsize=0.92\hsize}Y l >{\hsize=1.08\hsize}Y@{}}
\toprule
\textbf{Step} & \textbf{Operation} & \textbf{Methods} \\
\midrule
Remove a target & \texttt{remove} & General Editor\newline Removal Specialist \\
Recolor a target & \texttt{recolor} & General Editor\newline Masked Recolor \\
Restyle a target & \texttt{restyle} & General Editor\newline Masked Local Editor \\
Replace or add an object & \texttt{replace, add} & General Editor \\
Move a target & \texttt{move} & General Editor\newline Move Specialist \\
Resize a subject & \texttt{scale\_subject} & General Editor\newline Subject Scaling \\
\midrule
Whole-frame tone,\newline color or sharpness & \texttt{relight, generic} & General Editor\newline Photometric Operator \\
The same, in a target & \texttt{relight, generic} & General Editor\newline Local Photometric Operator \\
Other relighting\newline or harmonization & \texttt{relight, harmonize} & General Editor \\
\midrule
Named filter look & \texttt{filter} & General Editor\newline Style Filter \\
Named tone grade & \texttt{filter} & Style Filter \\
Named artist\newline (reference painting) & \texttt{filter} & General Editor\newline Neural Style Transfer\newline Multi-Reference Editor \\
Sketch, line art,\newline edge map & \texttt{filter} & General Editor\newline Neural Style Transfer \\
\midrule
Flip, or rotation by\newline a stated amount & \texttt{flip, rotate} & Flip and Exact Rotation \\
Rotation toward a\newline level or plumb goal & \texttt{rotate} & General Editor\newline Rotation Tool \\
Zoom in & \texttt{zoom\_in} & Zoom-In Crop \\
Zoom out or outpaint & \texttt{zoom\_out} & General Editor\newline Canvas Extension \\
Camera viewpoint change & \texttt{camera\_view} & General Editor \\
Any other request & \texttt{generic} & General Editor \\
\bottomrule
\end{tabularx}
\vspace{-1em}
\end{table}

\begin{table*}[t]
\centering
\caption{Methods of the capability pool, each with its backend, the claims established by code or by construction, and what is left to the judge. A candidate that breaks its guarantee is removed before judging, and the record of each remaining one carries both columns (\cref{sec:supp-records}).}
\label{tab:supp-pool}
\vspace{-0.5em}
\scriptsize
\setlength{\tabcolsep}{3pt}
\begin{tabularx}{\textwidth}{@{}P{0.155\textwidth} Y Y Y@{}}
\toprule
\textbf{Method} & \textbf{Backend} & \textbf{Established by Code or Construction} & \textbf{Left to the Judge} \\
\midrule
General Editor & The backbone (FLUX.2-klein-9B or Qwen-Image-Edit-2511) edits the whole frame. & None & Requested change, preservation, realism. \\
\midrule
Masked Local Editor & Target mask from grounding, SAM 2.1~\cite{ravi2025sam} and VLM review. The general editor edits the frame, and only its pixels inside the dilated and feathered mask are kept. & Unchanged outside the feathered mask (code). & Requested look in the mask, seam realism. \\
Masked Recolor & As above, with a tighter mask. & Unchanged outside the feathered mask (code). & Requested color, coverage, seam. \\
Removal Specialist & OSOR~\cite{zhou2026osor} with the reviewed target mask, whose output is used directly. & Canvas size unchanged (code). & Target and attached parts removed, background reconstructed, neighbors and supports kept. \\
Move Specialist & A diffusion transformer trained for object movement, conditioned on the source mask and a destination box designed by the VLM. Pixels outside the source and destination regions are restored from the source. & None & Object gone from its origin, same object at the destination, size consistent with depth, blending. \\
Subject Scaling & Reviewed mask, ViTMatte~\cite{yao2024vitmatte} alpha matte, uniform rescaling anchored at the feet, inpainting of the old footprint, contact shadow and color harmonization. A second candidate refines the composite with the general editor. & Every pixel outside the old and new subject and its shadow equals the source (code). No candidate whose scaled subject would cross a frame edge (construction). & Size change in the requested direction, recognizable subject, no halo. \\
\midrule
Photometric Operator & Closed-form brightness, contrast, saturation and color-temperature adjustments in CIELAB, and an unsharp mask for sharpening. & Canvas size unchanged, and every requested statistic moves in its direction (code). & Amount fits the wording, no other change. \\
Local Photometric Operator & The same, inside the feathered mask. & Unchanged outside the feathered mask (code). & As above, and the scope matches the wording. \\
\midrule
Style Filter & 25 deterministic color grades, photographic filters and drawing styles. & Canvas size unchanged (code). & The requested look. \\
Neural Style Transfer & Style transfer~\cite{gatys2016style} with VGG-19~\cite{simonyan2015vgg} features from a public-domain reference painting. & None & Style visible, content recognizable. \\
Multi-Reference Editor & The general editor with the reference painting as a second input image. & None & Style visible, content recognizable. \\
\midrule
Flip and Exact Rotation & Index reversal for flips and index permutation for rotations by multiples of 90$^\circ$. Other stated angles are resampled and cropped to the largest rectangle without empty corners. & For flips and multiples of 90$^\circ$, the output is the exact pixel permutation of the source (code). & Direction, and the amount for other angles. \\
Rotation Tool & For a level or plumb goal, the angle comes from the dominant long lines (length-weighted median of Hough segments within $25^\circ$ of level or plumb), and otherwise from the analyzer. FLUX.1-Fill~\cite{bfl2024flux1fill} generates a border ring, the extended frame is rotated at native scale, and the original window is cropped back. & Interior resampled from the source, generation confined to the border ring (construction). & Tilt direction and amount, corners. \\
Zoom-In Crop & A crop centered on the grounded target, upsampled with Swin2SR~\cite{conde2022swin2sr} and resized back. & The zoom factor exceeds 1 and the crop is smaller than the source (code). No editing model runs (construction). & Subject framing, upsampling artifacts. \\
Canvas Extension & The source is placed on a larger canvas whose border is filled by the general editor, with FLUX.1-Fill as a fallback. & The reported window places the source inside the wider frame (code). & Kept content, seam, plausible new area. \\
\bottomrule
\end{tabularx}
\vspace{-1em}
\end{table*}

\begin{table*}[t]
\centering
\caption{Measurements of the verification record, chosen by the operation of the step. Code computes them from the pixels in the same way for the source and every candidate. The change map and the close-up are images, and every other measurement is one line of text per image.}
\label{tab:supp-measurements}
\vspace{-0.5em}
\footnotesize
\setlength{\tabcolsep}{3pt}
\begin{tabularx}{\textwidth}{@{}P{0.14\textwidth} P{0.17\textwidth} Y Y@{}}
\toprule
\textbf{Measurement} & \textbf{Used For} & \textbf{Definition} & \textbf{Shown to the Judge} \\
\midrule
Visible Change & Every step & Share of pixels whose CIE76 color difference $\Delta E$ to the source exceeds 8, after a Gaussian blur with $\sigma = 1$ px of both images (no blur for detail edits). & One value per candidate, \eg ``visible change 4.2\% of the frame'' for a small local edit. \\
Change Map & Every step & Per-pixel $\Delta E$ to the source, shown as brightness on one scale shared by all candidates. & An image per candidate, black where a pixel is unchanged and brighter where it differs more. \\
Close-Up & Localized steps & The region that the request names, localized once in the source. The crop pads its box by 30\% of its size and covers at least 30\% of the frame on each axis. & A \texttt{BEFORE}\,|\,\texttt{AFTER} image per candidate, at the same place and scale for all. \\
Photometric Statistics & Tone, color and lighting & Brightness is the mean CIELAB $L^\ast/100$ and contrast the standard deviation of $L^\ast/100$. Saturation is the mean chroma $\sqrt{a^{\ast2}+b^{\ast2}}$. Clipped highlights and crushed shadows are the shares of pixels whose brightest channel is at least 250 or at most 5, and color shift is the change of mean $a^\ast$ and $b^\ast$. & One line for the source and one per candidate, \eg ``brightness 0.62, contrast 0.24, saturation 18.3, clipped 2.1\%, color shift $a^\ast$ $+0.3$ $b^\ast$ $+6.8$''. \\
Sharpness and Noise & Sharpening, denoising, blur & Variance of the Laplacian and the noise level of the luminance, with both images downsampled to a common size, and for a localized step also the sharpness inside the target relative to the source. & Ratios to the source, \eg ``sharpness $\times$1.84, noise $\times$0.71'' for a sharpened candidate. \\
Structure Kept & Global edits that\newline keep the layout & Mean luminance SSIM between the source and the candidate over the frame. & One value per candidate, where 1 means identical structure and lower values mean more change. \\
Orientation Match & Flips and rotations & Correlation of the candidate with the unchanged source and with the source mirrored left-right or top-bottom or rotated by 90$^\circ$, 180$^\circ$ or 270$^\circ$. & The correlations per candidate, from $-1$ to 1. \\
Framing & Zooming, canvas extension & Multi-scale template matching locates the candidate's view in the source (zoom in) or the source in the candidate (zoom out and extension). & Magnification and match score per candidate. \\
\bottomrule
\end{tabularx}
\vspace{-1em}
\end{table*}

\noindent \textbf{State.}
\method stores each session as a version graph. Every node holds one image state together with its parent, the step that produced it, its candidates, the judge's report, and whether it needs review. Accepted states are never modified. A later step adds a child node, and a user correction branches from the parent, so every accepted state remains available. The requirements that an accepted state satisfies, such as the requested change and the regions that had to stay, are carried into the prompts of the planner, the analyzer and the history check in later steps and turns.

\noindent \textbf{Turn.}
\Cref{alg:supp-turn} runs one user turn. The planner decides the number and order of the steps (\cref{lst:planner-system}), and each step starts from the state accepted by the previous step. When a step uses up its attempts, the re-planner either replaces the failed and pending steps or keeps the best attempt, which is then marked as needing review. After the turn, the history check verifies every earlier turn against the current image. On a confident violation, \method runs one repair turn and keeps its result only if the judge accepts it.

\noindent \textbf{Step.}
\Cref{alg:supp-step} runs one step. The analyzer rewrites the step into an explicit instruction for a mask-free editor, stating both the requested change and what must remain, sets the parameters of the applicable methods, and suggests how many candidates the step needs. The router selects the methods (\cref{tab:supp-routing}), each of which contributes at least one candidate. Every candidate receives a verification record, candidates that fail a code check are removed, and the judge compares the remaining candidates in a single call, in a shuffled order. An accepted candidate becomes the new state. Otherwise, the judge's feedback is passed back to the analyzer, which revises the step for the next attempt.

\begin{algorithm}[t]
\caption{One turn of \method. $\mathcal{G}$ is the version graph, $x_0$ the original image, $x_{1:t-1}$ the results of the earlier turns and $r_{1:t}$ the requests of turns 1 to $t$.}
\label{alg:supp-turn}
\small
\begin{algorithmic}[1]
\Require head state $x$, requests $r_{1:t}$, carried requirements $C$
\State $P \gets \textsc{Plan}(r_t, \mathcal{G}, C)$ \Comment{ordered steps}
\State $i \gets 1$;\ $m \gets 0$ \Comment{$m$ counts the re-plans}
\While{$i \le |P|$}
  \State $(o, y, F) \gets \textsc{Step}(P_i, x, C, \mathit{last} = [m = 2])$
  \If{$o = \texttt{exhausted}$} \Comment{only when $m < 2$}
    \State $m \gets m + 1$;\ $P' \gets \textsc{Replan}(r_t, P_{<i}, P_i, F, P_{>i}, C)$
    \If{$P' = \emptyset$} \Comment{keep the best attempt}
      \State $x \gets y$, marked needs review
      \State $C \gets \textsc{Carried}(x)$;\ $i \gets i + 1$
    \Else
      \State $P \gets P_{<i} \oplus P'$ \Comment{$P'_1$ inherits $F$}
    \EndIf
  \ElsIf{$o = \texttt{rollback}$}
    \State $i \gets i + 1$
  \Else \Comment{$o$ is \texttt{accepted} or \texttt{advanced}}
    \State $x \gets y$;\ $C \gets \textsc{Carried}(x)$;\ $i \gets i + 1$
  \EndIf
\EndWhile
\If{$t > 1$}
  \State $h \gets \textsc{History}(x_0, x_{1:t-1}, x, r_{1:t}, C)$
  \If{$h$ has a violation with confidence $\ge 0.4$ and a repair}
    \State $x \gets \textsc{Repair}(h.\mathit{repair}, x)$ \Comment{at most once}
    \State $h \gets \textsc{History}(x_0, x_{1:t-1}, x, r_{1:t}, C)$ \Comment{recorded}
  \EndIf
\EndIf
\State \Return $x$
\end{algorithmic}
\end{algorithm}

\begin{algorithm}[t]
\caption{One step of \method. $\pi$ is a shuffled presentation order and $y_{\mathrm{best}}$ the best candidate over all attempts.}
\label{alg:supp-step}
\small
\begin{algorithmic}[1]
\Require step $s$, source state $x$, carried requirements $C$, flag $\mathit{last}$
\State $F \gets$ feedback inherited from a replaced step, if any
\For{attempt $a = 0, 1, 2$}
  \State $q \gets \textsc{Analyze}(s, C, F)$ \Comment{instruction and parameters}
  \State $R \gets \textsc{Route}(q)$ \Comment{\cref{tab:supp-routing}}
  \State generate up to 3 new candidates, $\geq 1$ per method in $R$
  \State $Y \gets$ candidates that pass every code check, with records
  \If{$Y = \emptyset$}
    \State $F \gets F \cup \{\text{no candidate}\}$;\ \textbf{continue}
  \EndIf
  \State $J \gets \textsc{Judge}(q, x, \pi(Y))$ \Comment{one VLM call}
  \If{$J$ passes its best candidate $y^\ast$}
    \State \Return $(\texttt{accepted}, y^\ast, F)$
  \EndIf
  \State $F \gets F \cup \{J.\mathit{notes}\}$ \Comment{feedback for the next attempt}
\EndFor
\If{no attempt produced a candidate}
  \State \Return $(\texttt{rollback}, x, F)$
\ElsIf{$\mathit{last}$}
  \State \Return $(\texttt{advanced}, y_{\mathrm{best}}, F)$ \Comment{needs review}
\EndIf
\State \Return $(\texttt{exhausted}, y_{\mathrm{best}}, F)$
\end{algorithmic}
\end{algorithm}

\noindent \textbf{Acceptance.}
The judge returns one JSON object (\cref{lst:judge-system}) with a short note, a pass/fail verdict and a score from 1 to 10 for every candidate, followed by the best candidate and a complete ranking. Only the judge can accept a candidate. The step accepts the best candidate when the judge passes it, and otherwise passes the notes back to the analyzer as feedback for the next attempt.

\noindent \textbf{Robustness.}
A method that fails, for example because grounding finds no target, loses only its own candidate, and a step that never produced a candidate leaves the state unchanged. \method keeps the best candidate across the attempts of a step.

\section{Routing}
\label{sec:supp-routing}

\noindent \textbf{Routing Rules.}
The router is deterministic code (\cref*{sec:editing_methods} of the main paper). It turns the recommendation of the planner and analyzer, that is, the operation, target, executor and parameters of a step, into the methods of \cref{tab:supp-routing}, which \cref{tab:supp-pool} describes. Every masked method is raced against the general editor, so the judge can fall back to the general editor when a mask is imperfect.

\section{Verification Records}
\label{sec:supp-records}

\noindent \textbf{Definition.}
A verification record holds, for one candidate, the checks that code runs on its pixels (\cref*{sec:evidence_method} of the main paper). These are the share of visibly changed pixels, a change map, a close-up of the region that the request names, the measurements that the step calls for (\cref{tab:supp-measurements}), and the outcomes of code checks, which state what code has verified and what remains for visual judgment. Every record is computed in the same way from the source image and the candidate, and it names no method. What code cannot decide, such as whether an edit looks real, the judge assesses from the photographs (\cref{lst:judge-system}).

\noindent \textbf{Close-Up.}
For a localized step, the grounder names the region of the source image that the request refers to, such as ``the man's shirt'' or ``the sky behind the house'', and localizes it once, with a VLM box refined by SAM 2.1 and reviewed. Every candidate then gets a \texttt{BEFORE}\,|\,\texttt{AFTER} close-up of this same region, both halves at a height of 448 px, so that all candidates are inspected at the same place and scale, whichever method produced them.

\noindent \textbf{Change Map.}
The change map shows where and how strongly each candidate differs from the source, on one fixed scale for all candidates, so that the judge can compare the extent of the changes across candidates. It is not a photograph, and the judge does not assess color or realism from it.

\noindent \textbf{Measurements.}
The visible change counts the pixels whose CIELAB color differs from the source visibly ($\Delta E > 8$), after a light blur that suppresses compression noise. For sharpening, deblurring, denoising and blur requests, no blur is applied, since these edits change exactly the fine detail that a blur would hide. The other measurements depend on the operation of the step (\cref{tab:supp-measurements}), such as the brightness, clipping and color shift in \cref*{fig:evidence}(b) of the main paper. 

\noindent \textbf{Code Checks.}
Each method also declares what it guarantees by construction or verifies by code, such as unchanged pixels outside its mask or an exact pixel permutation for a flip (\cref{tab:supp-pool}). A candidate that fails such a check is removed before judging. The record of every remaining candidate lists the checks it passed under \texttt{VERIFIED} and what code cannot decide under \texttt{TO JUDGE}.

\noindent \textbf{Presentation.}
The judge receives one message with the instructions of \cref{lst:judge-system}, the request of the step with its acceptance criteria and regions to keep, whether the step is localized or global, the images and the text of the records. The images are \texttt{BEFORE} and then each candidate \texttt{TRIAL n} with its change map and, for a localized step, its close-up. The text gives, for each candidate, the visible change, the measurements and the \texttt{VERIFIED} and \texttt{TO JUDGE} lines. For photometric steps, the line of the source comes first, and a short definition of each number follows.

\section{Benchmark Experiments}
\label{sec:supp-benchmarks}

\noindent \textbf{Backbones.}
Each backbone uses the same checkpoint and sampler with and without \method. Qwen-Image-Edit-2511 runs 40 steps with CFG 4 in BF16, and FLUX.2-klein-9B uses its distilled 4-step sampler. Within \method, the backbone is the general editor. It therefore also runs inside the masked methods, the refinement and harmonization of Subject Scaling, the Multi-Reference Editor and the fill of Canvas Extension, which all call the general editor. OSOR, FLUX.1-Fill for the rotation ring and the outpainting fallback, Swin2SR, neural style transfer and the closed-form operators are the same for all backbones. The prompts are used unchanged with every backbone.

\noindent \textbf{Inputs.}
\method and the standalone backbones receive the same original benchmark instructions, \method plans internally, and all scoring uses the original benchmark annotations. Standalone backbones use seed 42, and seeds 42, 43 and 44 for the three EdiVal turns. 

\noindent \textbf{Scoring.}
EdiVal is scored with its released code and its released evaluators GroundingDINO~\cite{liu2024groundingdino}, Qwen2-VL-7B~\cite{wang2024qwen2vl} and DINOv3~\cite{simeoni2025dinov3}, all unchanged. Its instruction following is cumulative, so a session passes turn $t$ only if the instruction of every turn up to $t$ is fulfilled by the output of that turn. EdiVal-O is the geometric mean of instruction following and content consistency. On Complex-Edit~\cite{yang2025complexedit}, Qwen3.8-27B replaces GPT-4o in the official prompts. Instruction following and identity preservation are rated from the source and the output with the official rubric, and perceptual quality from the output alone, each averaged over five samples on a scale from 0 to 10. The overall score is the mean of the three dimensions. The 8,496 requests comprise 531 instruction chains in each of the real and synthetic splits, at eight complexity levels. On ComplexBench-Edit~\cite{wang2025complexbenchedit}, Qwen3.8-27B replaces Gemini 2.0 Flash Thinking and rates each atomic instruction from the source and the output with the official rubric of its edit type. With atomic scores $a$, $b$ and $c$ in $[0,1]$, parallel requests score $(a+b+c)/3$, two-step chains $(ab+c)/2$ and three-step chains $abc$, reported $\times 100$. The reported average is the mean over its five groups of requests in \cref*{tab:complex-main}, from single-object edits to three-step chains.

\noindent \textbf{Agentic Baselines.}
\Cref{tab:supp-agents} lists how each agent of \cref*{tab:edival-main-agent} is run. All agents edit through one service that runs the standalone backbone with its recipe, so they differ only in their policy. A masked or inpainting tool becomes a backbone edit composited inside a SAM 2.1 mask. Each agent receives the original EdiVal instruction of the current turn and edits its own previous output. Attempt $a$ of turn $t$ uses seed $42 + 100a + t - 1$, so a first attempt uses the seed of the standalone backbone. The standalone rows of \cref*{tab:edival-main,tab:edival-main-agent} come from the same service, so all agents are compared against one standalone result of each backbone.

\begin{table}[t]
\centering
\caption{How each agentic baseline of \cref*{tab:edival-main-agent} of the main paper is run. Every agent edits with the same backbone and uses Qwen3.8-27B wherever it calls a multimodal LLM. $^\dagger$Re-implemented.}
\label{tab:supp-agents}
\vspace{-0.5em}
\small
\setlength{\tabcolsep}{3pt}
\begin{tabularx}{\columnwidth}{@{}l Y@{}}
\toprule
\textbf{Agent} & \textbf{How It Is Run} \\
\midrule
GenArtist$^\dagger$ & Official prompts. The unreleased editing loop is re-implemented, and its tools become backbone edits, masked for local tools. \\
MIRA$^\dagger$ & Trained policy model not released. The planning loop of the paper runs zero-shot, with Qwen3.8-27B serving as its policy model. \\
IMAGAgent & Released code and prompts, with closed models replaced by Qwen3.8-27B and the backbone, and inpainting by masked edits. \\
EditThinker-8B & Released code and 8B thinker. \\
VisionDirector-TF$^\dagger$ & Trained planner not released. A training-free loop uses the released verifier prompt and runs local goals as masked edits. \\
RISE-Agent & Released code, with Qwen3.8-27B as planner and verifier and without web search. \\
\makecell[l]{EditThinker\\(Qwen3.8-27B)} & Released code in its training-free mode, with Qwen3.8-27B as the thinker. \\
\bottomrule
\end{tabularx}
\vspace{-1em}
\end{table}

\noindent \textbf{Alternative Internal Judge.}
With Gemma-4-31B as the internal judge, \method improves every backbone on all three benchmarks (\cref{tab:supp-gemma}).

\begin{table}[t]
\centering
\caption{\method with Qwen3.8-27B or Gemma-4-31B as its internal judge, under unchanged external scoring. The columns are EdiVal-O at turns 1 / 2 / 3, the ComplexBench-Edit average (CBE) and the Complex-Edit overall score (CE-O).}
\label{tab:supp-gemma}
\vspace{-0.5em}
\footnotesize
\setlength{\tabcolsep}{3pt}
\begin{tabular}{lccc}
\toprule
\textbf{Model} & \textbf{EdiVal-O} & \textbf{CBE} & \textbf{CE-O} \\
\midrule
Qwen-Image-Edit-2511 & 75.97 / 58.58 / 47.64 & 54.40 & 7.371 \\
\quad + \method (Qwen3.8-27B) & 80.64 / 67.02 / 55.46 & 60.83 & 7.445 \\
\quad + \method (Gemma-4-31B) & 80.34 / 66.28 / 54.17 & 60.53 & 7.457 \\
\midrule
FLUX.2-klein-9B & 81.37 / 66.30 / 53.20 & 61.45 & 7.677 \\
\quad + \method (Qwen3.8-27B) & 84.46 / 71.10 / 59.45 & 64.51 & 7.713 \\
\quad + \method (Gemma-4-31B) & 85.17 / 71.64 / 60.25 & 63.44 & 7.715 \\
\bottomrule
\end{tabular}
\vspace{-1em}
\end{table}

\noindent \textbf{Route Usage.}
Specialized methods deliver 19\% to 31\% of the accepted steps, depending on the backbone and the benchmark (\cref{tab:route-usage-supp}). Masked edits and object removals are the most frequent specialist wins.

\begin{table*}[t]
\centering
\caption{Accepted steps by method for each backbone and benchmark (\% of the accepted steps on the final edit chain of each request). Masked Local Editor and Masked Recolor run the backbone inside a mask. Removal Specialist, Photometric Operators (whole-frame and local) and Style Filter do not use the backbone. Zoom and Other covers Zoom-In Crop, Canvas Extension, Flip and Exact Rotation, Rotation Tool and the two style-transfer methods.}
\label{tab:route-usage-supp}
\vspace{-0.5em}
\footnotesize
\setlength{\tabcolsep}{3pt}
\begin{tabular}{ll*{7}{c}r}
\toprule
\textbf{Benchmark} & \textbf{Backbone} & \textbf{\shortstack{Masked Local\\Editor}} & \textbf{\shortstack{Masked\\Recolor}} & \textbf{\shortstack{Removal\\Specialist}} & \textbf{\shortstack{Photometric\\Operators}} & \textbf{\shortstack{Style\\Filter}} & \textbf{\shortstack{Zoom and\\Other}} & \textbf{\shortstack{General\\Editor}} & \textbf{\shortstack{Accepted\\Steps}} \\
\midrule
\multirow{2}{*}{EdiVal} & Qwen-Image-Edit-2511 & 9.2 & 6.7 & 8.1 & 0.0 & 0.0 & 0.6 & 75.4 & 1,253 \\
 & FLUX.2-klein-9B & 7.8 & 7.7 & 3.3 & 0.0 & 0.0 & 0.0 & 81.2 & 1,628 \\
\midrule
\multirow{2}{*}{ComplexBench-Edit} & Qwen-Image-Edit-2511 & 10.1 & 5.2 & 16.0 & 0.0 & 0.1 & 0.0 & 68.6 & 1,891 \\
 & FLUX.2-klein-9B & 8.3 & 5.3 & 7.2 & 0.0 & 0.0 & 0.0 & 79.2 & 2,246 \\
\midrule
\multirow{2}{*}{Complex-Edit} & Qwen-Image-Edit-2511 & 6.0 & 4.6 & 3.7 & 5.4 & 4.4 & 1.2 & 74.8 & 28,712 \\
 & FLUX.2-klein-9B & 4.8 & 4.7 & 1.8 & 4.5 & 3.6 & 1.1 & 79.5 & 33,224 \\
\bottomrule
\end{tabular}
\vspace{-1em}
\end{table*}

\section{Data Engine}
\label{sec:supp-data}

\noindent \textbf{Sources and Requests.}
Source images are real photographs from the training split of Open Images~\cite{kuznetsova2020openimages}. Qwen3.8-27B writes the multi-turn requests. It screens each photograph for readable text, separable objects, grayscale content and people, assigns it a session of 3 to 5 distinct edit types from the 30 types of \cref{tab:supp-edit-types} that the photograph supports, writes one instruction per type, turn by turn, each referring to the result of the previous turn, and checks each session against its photograph.

\noindent \textbf{Generation and Filtering.}
\method executes each session turn by turn with the settings of \cref{tab:supp-settings} and FLUX.2-klein-9B as the general editor. Qwen3.8-27B serves as planner, analyzer, grounder and judge. Of the 40,726 executed sessions (163,806 turns), the judge accepts 91.4\% of the turns (\cref{tab:supp-data-stats}). Because a session is kept only if the judge accepts every one of its turns, 28,673 sessions (70.4\%) pass this filter, and no kept session contains a result marked as needing review. After we also remove the 591 sessions that are both in the lowest 15\% of quality and average more than two attempts per turn, the remaining 28,082 sessions, 69.0\% of those executed, contain 111,755 turns.

\begin{table}[t]
\centering
\caption{Data engine statistics by edit group over all 163,806 executed turns. Accepted is the share of turns accepted by the judge, and First the share whose steps are all accepted at the first attempt, without re-planning. Groups follow \cref{tab:supp-edit-types}. Attempts and candidates are means per turn.}
\label{tab:supp-data-stats}
\vspace{-0.5em}
\small
\setlength{\tabcolsep}{2.5pt}
\begin{tabular}{@{}lrrrrr@{}}
\toprule
\textbf{Group} & \textbf{Turns} & \textbf{\shortstack{Accepted\\(\%)}} & \textbf{\shortstack{First\\(\%)}} & \textbf{Attempts} & \textbf{\shortstack{Candi-\\dates}} \\
\midrule
Content & 46,675 & 94.1 & 79.6 & 1.58 & 4.06 \\
Style \& Effects & 22,864 & 94.6 & 82.0 & 1.58 & 4.03 \\
Scene & 28,595 & 92.2 & 70.4 & 2.23 & 5.15 \\
Photometric & 22,841 & 93.8 & 65.5 & 2.13 & 3.35 \\
Geometry & 31,390 & 83.3 & 69.6 & 2.56 & 5.29 \\
Enhancement & 11,441 & 89.9 & 63.8 & 2.59 & 4.63 \\
\midrule
All & 163,806 & 91.4 & 73.3 & 2.03 & 4.42 \\
\bottomrule
\end{tabular}
\vspace{-1em}
\end{table}

\noindent \textbf{Compute.}
Generation runs on nodes of 8 H200 GPUs. With speculative decoding (DFlash2~\cite{inco2026dflash2}), the 40,726 sessions take about 400 GPU-hours. Every 10,000 sessions take about 100 GPU-hours, and about 70\% of them are kept as high quality.

\noindent \textbf{Training Pairs.}
Each kept session gives one training pair, made of the source image, one instruction for the whole session and the final image. A pair ends before the first turn that changes the image size and needs at least two applied edits. This gives 26,251 pairs with 3.9 edits per pair on average. Qwen3.8-27B composes the instruction from those of the delivered steps, resolving the final intent by dropping overridden or cancelled edits. The pairs are split by source image into 25,753 training and 498 validation pairs. The same pairs train the students for all three benchmarks.

\noindent \textbf{Baseline Data.}
For the matched baseline, FLUX.2-klein-9B edits every turn of the same sessions directly with the original per-turn instructions, each turn on its own previous output, without planning or judging. Its pairs use the same kept sessions, split and instructions as the \method pairs, with the final image taken from the FLUX.2-klein-9B chain, so the two datasets differ only in their images.

\noindent \textbf{Finetuning.}
We finetune all transformer weights of FLUX.2-klein-base-9B (9.1B parameters) and Qwen-Image-Edit-2511 (20.4B), with the text encoder and VAE frozen, using DiffSynth-Studio~\cite{diffsynth} with DeepSpeed ZeRO-3~\cite{rajbhandari2020zero} in BF16 on 8 H200 GPUs. Training uses a global batch of 8 (one per GPU), AdamW~\cite{loshchilov2019adamw} with a learning rate of $10^{-5}$ for FLUX.2-klein-base-9B and $2 \times 10^{-6}$ for Qwen-Image-Edit-2511 (after a five-update warm-up at one third of the rate), weight decay of 0.01 and 0.05, respectively, gradient clipping at 1.0, gradient checkpointing, images of up to $1024^2$ px at their native aspect ratio, and a fixed data order. Each model is trained for one epoch (3,220 steps). For evaluation, FLUX.2-klein-base-9B is sampled with 50 steps and CFG 4, and Qwen-Image-Edit-2511 with 40 steps and CFG 4. On EdiVal, the students and their untuned baselines receive rewritten instructions (\cref*{tab:data-engine}).

\begin{figure*}[p]
\centering
\includegraphics[width=\textwidth]{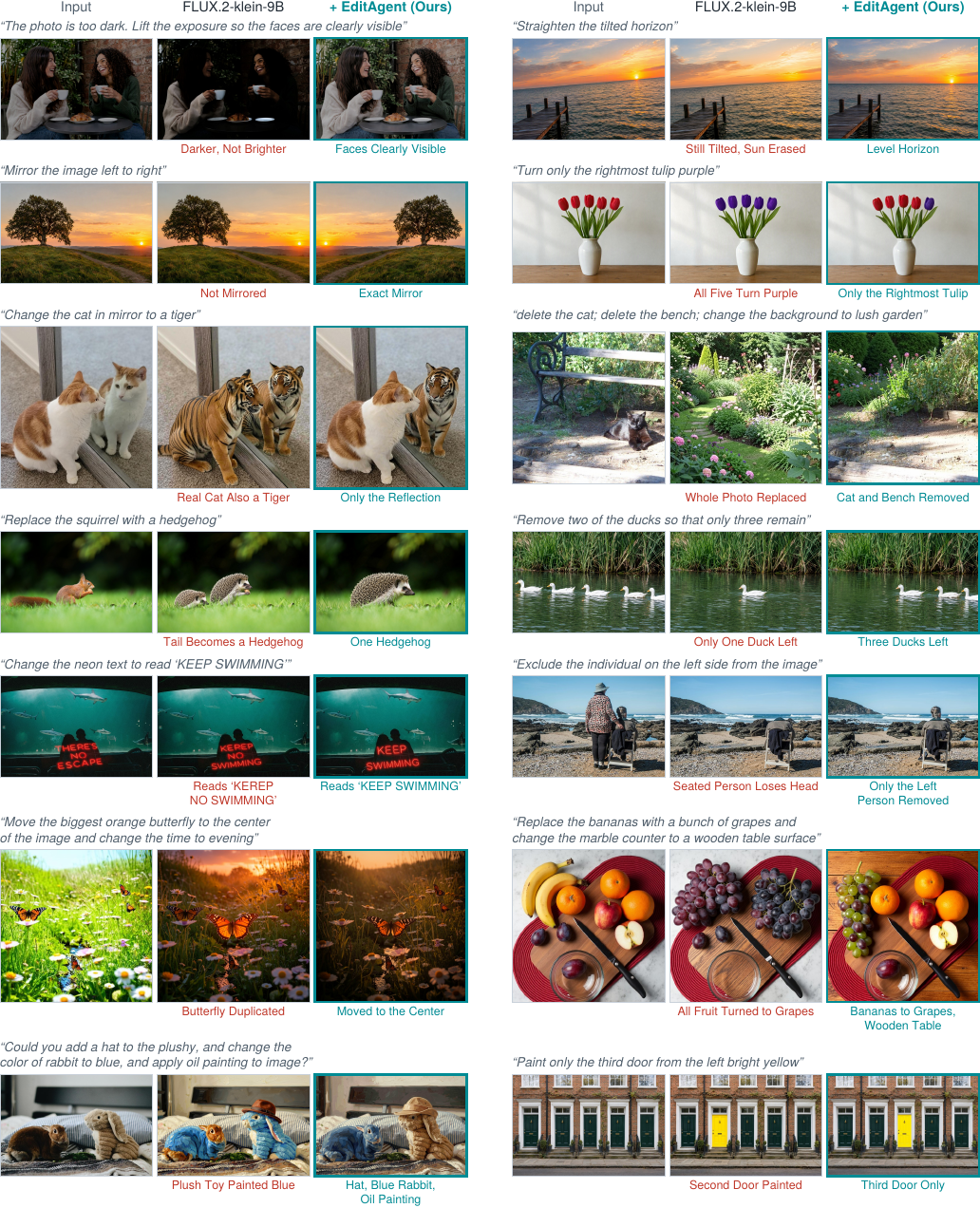}
\caption{Single-turn editing comparison with FLUX.2-klein-9B.}
\label{fig:supp-qual-single-flux}
\end{figure*}

\FloatBarrier
\twocolumn[{%
\section{Additional Qualitative Results}
\label{sec:supp-qualitative}
\Crefrange{fig:supp-qual-multi-qie}{fig:supp-qual-single-flux} compare more multi-turn and single-turn edits of \method with the backbone alone. For every turn, each row shows the instruction, the output of the backbone and the output of \method with the same backbone (teal frame). Red labels name the failure of the backbone output.\par
\vspace{1em}
\centering
\includegraphics[width=\textwidth]{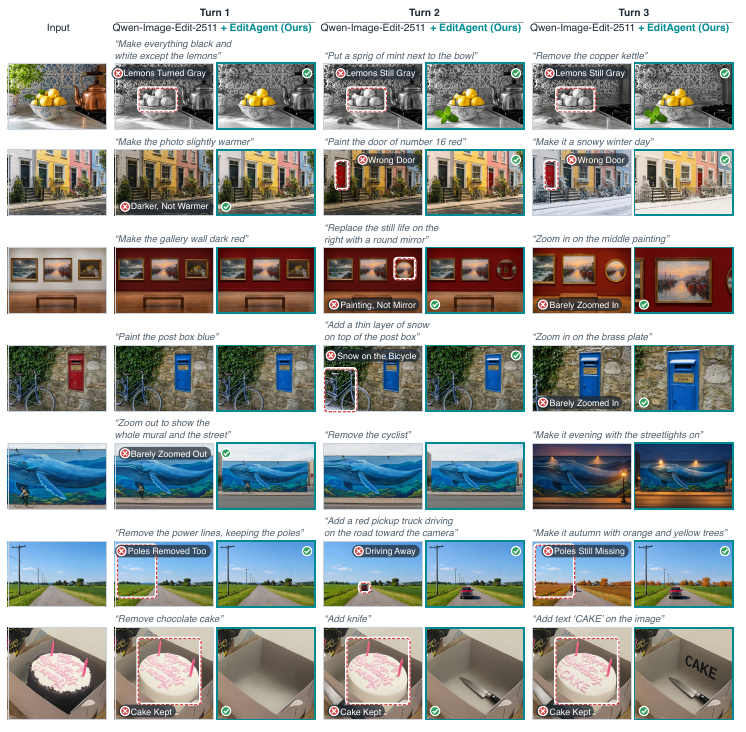}
\captionof{figure}{Multi-turn editing comparison with Qwen-Image-Edit-2511.}
\label{fig:supp-qual-multi-qie}
\vspace{1em}
}]

\begin{figure*}[p]
\centering
\includegraphics[width=\textwidth]{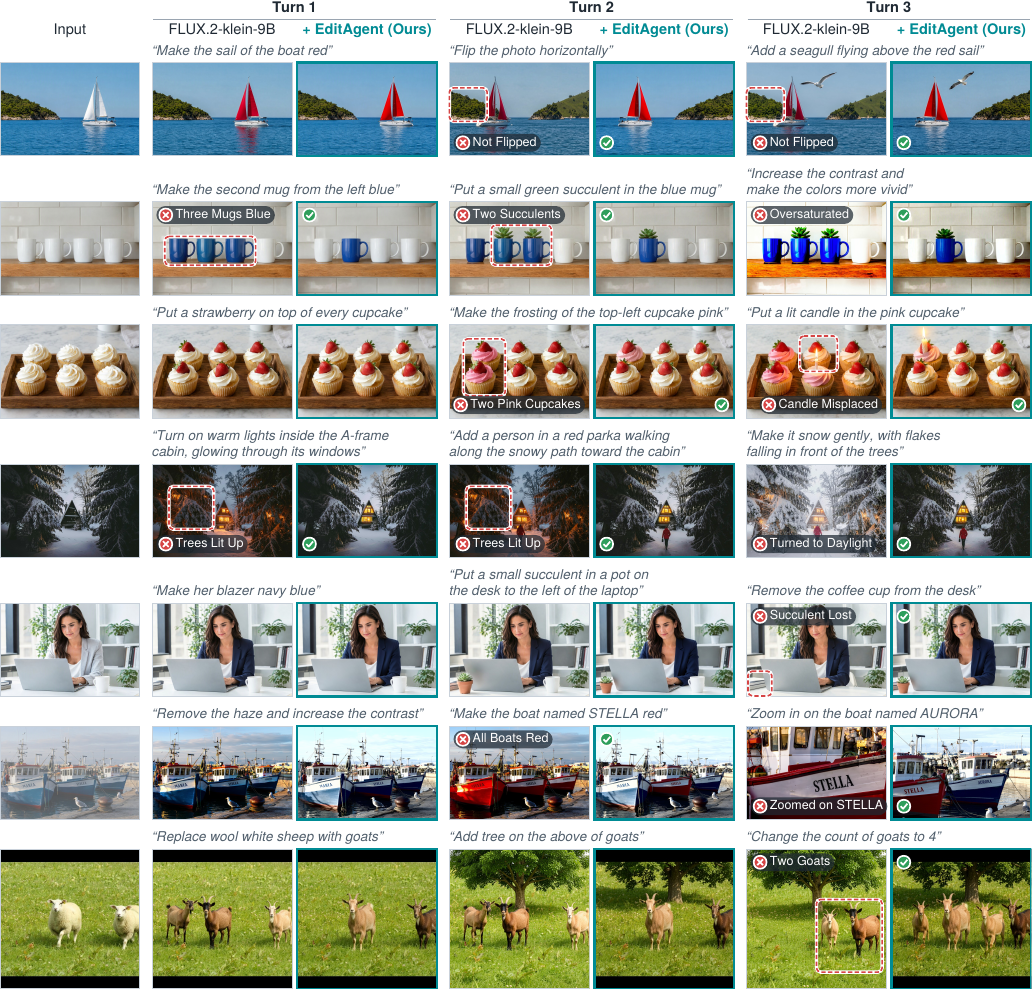}
\caption{Multi-turn editing comparison with FLUX.2-klein-9B.}
\label{fig:supp-qual-multi-flux}
\end{figure*}

\begin{figure*}[p]
\centering
\includegraphics[width=\textwidth]{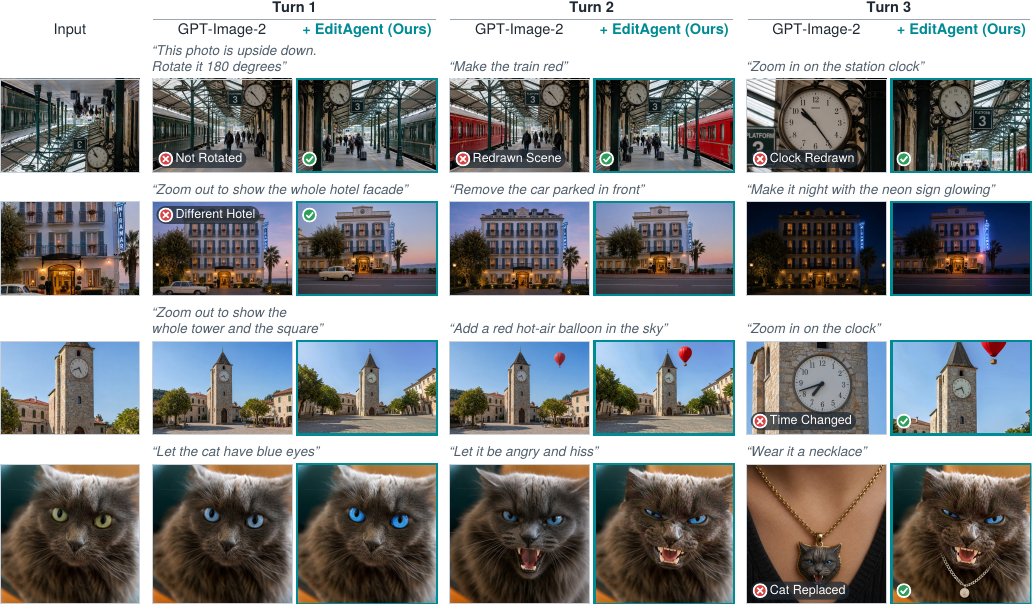}
\caption{Multi-turn editing comparison with GPT-Image-2.}
\label{fig:supp-qual-multi-gpt}
\vspace{1.2em}
\includegraphics[width=\textwidth]{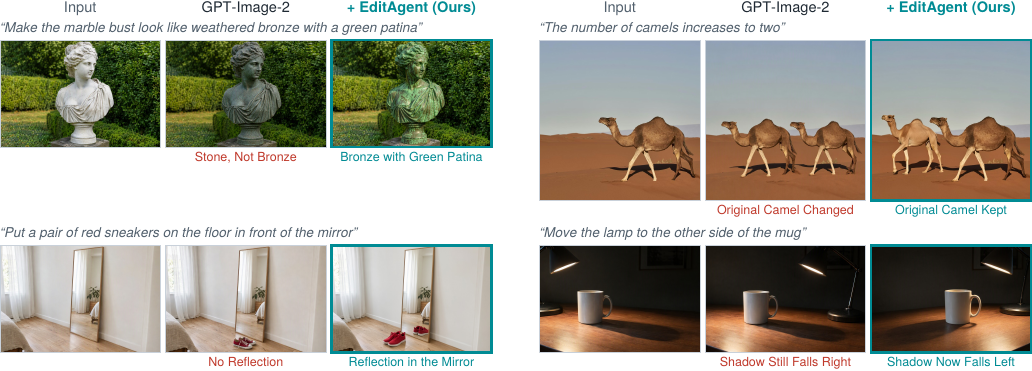}
\caption{Single-turn editing comparison with GPT-Image-2.}
\label{fig:supp-qual-single-gpt}
\end{figure*}

\begin{figure*}[p]
\centering
\includegraphics[width=\textwidth]{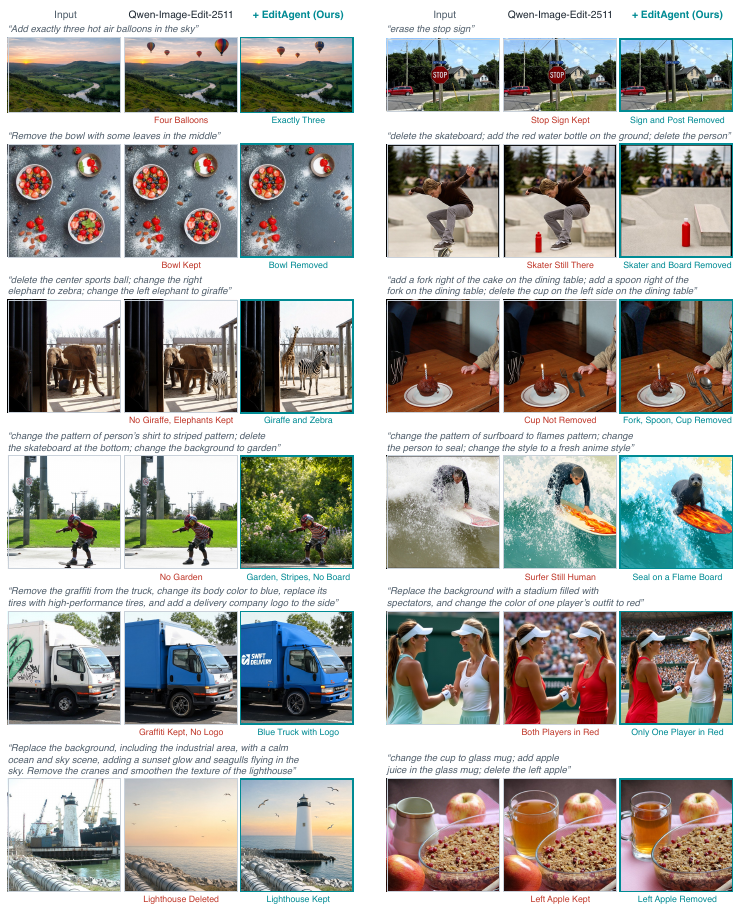}
\caption{Single-turn editing comparison with Qwen-Image-Edit-2511.}
\label{fig:supp-qual-single-qie}
\end{figure*}

\FloatBarrier
\section{Prompts}
\label{sec:supp-prompts}

This section gives the instructions (system prompts) of \method. In the listings, text in angle brackets, $\langle\cdot\rangle$, is filled in at run time. This includes the shared rules of \cref{lst:routing}, which the planner, analyzer and re-planner all receive and which are printed once. In the JSON schemas, \texttt{<\,\ldots\,>} describes a value that the model writes. \Cref{tab:supp-glossary} gives the names that the prompts use for terms of the paper.

\begin{table}[t]
\centering
\caption{Terms of the paper and their names in the prompts.}
\label{tab:supp-glossary}
\vspace{-0.5em}
\footnotesize
\begin{tabularx}{\linewidth}{@{}l Y@{}}
\toprule
\textbf{Term in the Paper} & \textbf{Name in the Prompts} \\
\midrule
Candidate & \texttt{TRIAL n} \\
Source Image of a Step & \texttt{BEFORE} \\
Change Map & \texttt{CHANGE-MAP} \\
Close-Up & \texttt{CLOSE-UP}, \texttt{THEN | NOW} close-up \\
General Editor & \texttt{edit\_model} \\
Masked Methods & \texttt{masked\_composite} \\
Target & \texttt{target\_desc} \\
Regions to Keep & \texttt{protect\_desc} \\
Carried Requirements & \texttt{carried\_constraints}, \texttt{derived\_constraints} \\
\bottomrule
\end{tabularx}
\end{table}

\medskip
\Needspace{13\baselineskip}
\noindent \textbf{Planner.}

\promptlisting{supp/prompts/planner_system.txt}{Planner instructions (system), one call per turn.}{lst:planner-system}

\promptlisting{supp/prompts/routing.txt}{Rules shared by the planner, analyzer and re-planner, covering the executor, the operations and their parameters, the editing methods of the capability pool and preservation.}{lst:routing}

\Needspace{13\baselineskip}
\noindent \textbf{Analyzer.}

\promptlisting{supp/prompts/analyzer_system.txt}{Analyzer instructions (system). The analyzer turns a planned step into an executable instruction with acceptance criteria and recommends the methods for its candidates.}{lst:analyzer-system}

\Needspace{13\baselineskip}
\noindent \textbf{Grounding.}

\promptlisting{supp/prompts/ground.txt}{Grounding prompt. Code converts the boxes to pixels, and SAM 2.1 refines them into a mask.}{lst:ground}

\promptlisting{supp/prompts/mask_review.txt}{Mask review, with the original image, the image with a cyan outline of the mask and the binary mask. A rejected mask is extended with the named missing parts or replaced by the corrected boxes, then reviewed again.}{lst:mask-review}

\Needspace{13\baselineskip}
\noindent \textbf{Judge.}

\promptlisting{supp/prompts/judge_system.txt}{Judge instructions (system) for one step. The candidates follow in a shuffled order, each with its verification record.}{lst:judge-system}

\Needspace{13\baselineskip}
\noindent \textbf{Re-Planner.}

\promptlisting{supp/prompts/replanner_system.txt}{Re-planner instructions (system), used when every attempt of a step has failed, to plan the rest of the turn.}{lst:replanner-system}

\Needspace{13\baselineskip}
\noindent \textbf{History Check.}

\promptlisting{supp/prompts/history_system.txt}{History-check instructions (system), run after every turn from turn 2 on to verify all earlier turns.}{lst:history-system}

\end{document}